\documentclass[11pt]{article}

\usepackage[preprint]{acl}

\usepackage{times}
\usepackage{latexsym}

\usepackage[T1]{fontenc}

\usepackage[utf8]{inputenc}

\usepackage{microtype}

\usepackage{inconsolata}

\usepackage{graphicx}
\usepackage{tcolorbox}

\title{KG-MuLQA: A Framework for KG-based Multi-Level QA Extraction \\ and Long-Context LLM Evaluation}

\author{Nikita Tatarinov,\textsuperscript{\Letter} Vidhyakshaya Kannan,\textsuperscript{*} Haricharana Srinivasa,\textsuperscript{*} Arnav Raj, \\ {\bf Harpreet Singh Anand, Varun Singh, Aditya Luthra, Ravij Lade,} \\
  {\bf Agam Shah,\textsuperscript{\Letter} Sudheer Chava} \\\\
  Georgia Institute of Technology \\
  \Letter\;Corresponding Authors: \{ntatarinov3, ashah482\}@gatech.edu \\
  \textsuperscript{*} Indicates equal contribution
}

\usepackage{booktabs} 
\usepackage{marvosym} 
\usepackage{pifont} 
\usepackage{enumitem} 
\usepackage{subcaption} 
\usepackage{listings} 
\usepackage[table]{xcolor} 
\usepackage{amssymb} 
\usepackage{arydshln} 

\newcommand\frameworkshortname{{\sc KG-MuLQA}}
\newcommand\frameworkshortnamebold{{\sc \textbf{KG-MuLQA}}}
\newcommand\fullname{\textbf{K}nowledge-\textbf{G}raph-based \textbf{Mu}lti-\textbf{L}evel \textbf{Q}uestion-\textbf{A}nswer Extraction}

\newcommand\datasetshortname{{\sc KG-MuLQA-D}}
\newcommand\datasetshortnamebold{{\sc \textbf{KG-MuLQA-D}}}

\lstdefinelanguage{SPARQL}{
  morekeywords={SELECT, WHERE, FILTER, OPTIONAL, UNION, GRAPH, PREFIX, DISTINCT, LIMIT, ORDER, BY, ASC, DESC},
  sensitive=false,
  morecomment=[l][\color{gray}]{\#},
  morestring=[b]"
}

\begin{document}
\maketitle
\begin{abstract}
We introduce {\sc KG-MuLQA} (\textbf{K}nowledge-\textbf{G}raph-based \textbf{Mu}lti-\textbf{L}evel \textbf{Q}uestion-\textbf{A}nswer Extraction): a framework that (1) extracts QA pairs at multiple complexity levels (2) along three key dimensions -- multi-hop retrieval, set operations, and answer plurality, (3) by leveraging knowledge-graph-based document representations. This approach enables fine-grained assessment of model performance across controlled difficulty levels. Using this framework, we construct a dataset of 20,139 QA pairs based on financial credit agreements and evaluate 16 proprietary and open-weight Large Language Models, observing that even the best-performing models struggle with set-based comparisons and multi-hop reasoning over long contexts. Our analysis reveals systematic failure modes tied to semantic misinterpretation and inability to handle implicit relations.
\end{abstract}

\section{Introduction}
\label{sec:introduction}

\begin{figure}[t]
    \centering
    \includegraphics[width=\linewidth]{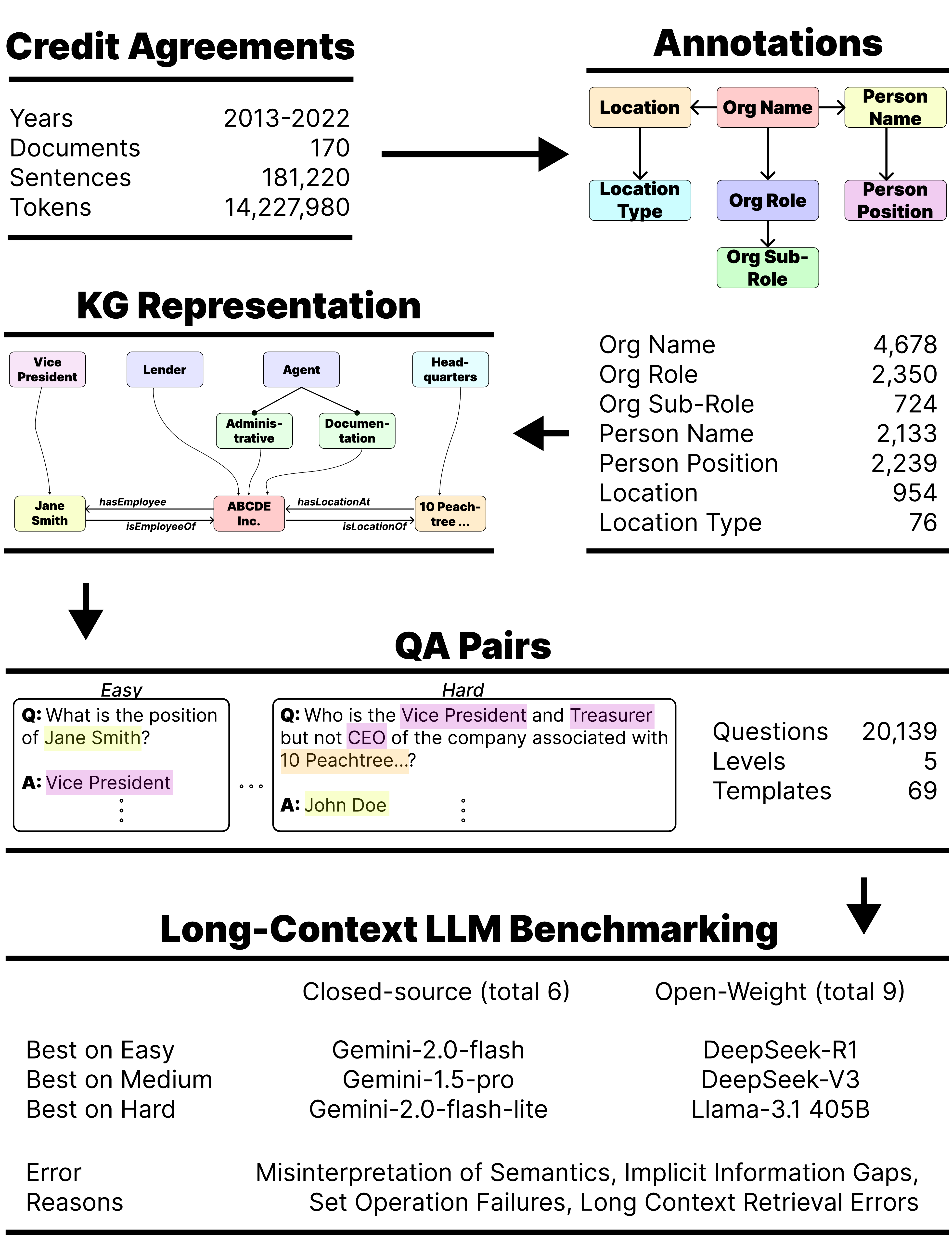}
    \caption{Overview of \frameworkshortname{}. Credit agreements are annotated to identify entities and their relationships, forming a knowledge graph representation. This graph is then used to \textit{systematically extract multi-level QA pairs}, which serve as the basis for benchmarking long-context LLMs.}
    \label{fig:our_pipeline}
\end{figure}

\begin{table*}[t]
    \caption{Comparison of \datasetshortname{} with existing long-context LLM benchmarks. \textbf{\textit{Multi-hop Reasoning}} marks whether the dataset requires reasoning across multiple evidence pieces; \datasetshortname{} explicitly defines and scales multi-hop paths through its knowledge-graph structure, enabling controlled evaluation of relational inference depth. \textbf{\textit{Systematic Q-Extraction}} denotes whether questions are generated deterministically rather than manually or by LLMs; \datasetshortname{} automates QA generation from annotated graphs, preserving real document semantics without synthetic injection. \textbf{\textit{Question Complexity}} shows whether the benchmark categorizes questions by difficulty; \datasetshortname{} introduces a principled, multi-dimensional notion of complexity (hops + set operations + plurality), allowing granular performance analysis across reasoning levels.}
    \label{tab:literature_review}

    \renewcommand{\arraystretch}{1.1}  
    \centering
    \resizebox{\textwidth}{!}{%
    \begin{tabular}{@{}lcccccc@{}}
        \toprule
         & \textbf{Long} & \textbf{Multi-hop} & \textbf{Systematic} & \textbf{Question} & & \\
        \textbf{Dataset} & \textbf{context} & \textbf{reasoning} & \textbf{Q-extraction} & \textbf{complexity} & \textbf{Document Type} & \# \textbf{Samples} \\
        \midrule
        L-Eval \citep{an-etal-2024-l} & \ding{52} & \ding{56} & \ding{56} & \ding{56} & Earnings Calls + Misc & 2,008 \\
        CLongEval \citep{qiu-etal-2024-clongeval} & \ding{52} & partially & \ding{56} & \ding{56} & THUnews + Misc & 7,267 \\
        Marathon \citep{zhang-etal-2024-marathon} & \ding{52} & partially & \ding{56} & \ding{56} & Prior Datasets & 1,530 \\
        MMLongBench-DOC \citep{ma2024mmlongbenchdocbenchmarkinglongcontextdocument} & \ding{52} & partially & \ding{56} & \ding{56} & arXiv + Misc & 1,082 \\
        DocFinQA \citep{reddy-etal-2024-docfinqa} & \ding{52} & partially & \ding{56} & \ding{56} & SEC 10-K \& 10-Q & 7,437 \\
        NOCHA \citep{karpinska-etal-2024-one} & \ding{52} & \ding{52} & \ding{56} & \ding{56} & Fictional Books & 2,002 \\
        LooGLE \citep{li-etal-2024-loogle} & \ding{52} & \ding{52} & \ding{56} & \ding{56} & arXiv + Misc & 6,448 \\
        BAMBOO \citep{dong-etal-2024-bamboo} & \ding{52} & \ding{52} & \ding{56} & \ding{56} & ACL Anthology + Misc & 1,502 \\
        LongBench \citep{bai-etal-2024-longbench} & \ding{52} & \ding{52} & \ding{56} & \ding{56} & Previous Datasets & 4,750 \\
        Loong \citep{wang-etal-2024-leave} & \ding{52} & \ding{52} & \ding{56} & \ding{56} & SEC + Misc & 1,600 \\
        FinDVer \citep{zhao-etal-2024-findver} & \ding{52} & \ding{52} & \ding{56} & \ding{56} & SEC 10-K \& 10-Q & 2,100 \\
        M4LE \citep{kwan-etal-2024-m4le} & \ding{52} & \ding{56} & partially & \ding{56} & Prior Datasets & unspecified \\
        Michelangelo \citep{vodrahalli2024michelangelolongcontextevaluations} & \ding{52} & partially & partially & \ding{56} & synthetic & unspecified \\
        \( \infty \)Bench \citep{zhang-etal-2024-bench} & \ding{52} & partially & needles & \ding{56} & Novels + Misc & 3,946 \\
        mLongRR \citep{agrawal-etal-2024-evaluating} & \ding{52} & \ding{52} & needles & \ding{56} & BBC News & unspecified \\
        \citep{gupta-etal-2024-systematic} & \ding{52} & \ding{56} & \ding{56} & partially & Financial News & 560 \\
        DocMath-Eval \citep{zhao-etal-2024-docmath} & \ding{52} & partially & \ding{56} & partially & SEC 10-K \& 10-Q & 4,000 \\
        FanOutQA \citep{zhu-etal-2024-fanoutqa} & \ding{52} & \ding{52} & \ding{56} & partially & Wikipedia & 7,305 \\
        RULER \citep{hsieh2024rulerwhatsrealcontext} & \ding{52} & \ding{52} & needles & partially & SEC 10-K \& 10-Q & 30,000 \\
        BABILong \citep{kuratov2024babilongtestinglimitsllms} & \ding{52} & \ding{52} & partially & partially & Books + Misc & 60,000 \\
        MuSiQue \citep{trivedi-etal-2022-musique} & \ding{56} & \ding{52} & \ding{52} & \ding{52} & Wikipedia & 24,814 \\
        \midrule
        \datasetshortnamebold{} (ours) & \ding{52} & \ding{52} & \ding{52} & \ding{52} & SEC Credit Agreements & 20,139 \\
        \bottomrule
    \end{tabular}
    }
\end{table*}

The increasing context length of recent Large Language Models (LLMs) has led to growing interest in evaluating their capabilities. Researchers have studied challenges like the \textit{Lost in the Middle} problem \citep{liu-etal-2024-lost}, examined the limiting factors of scaling models to long context such as the constraints imposed by RoPE’s base value \citep{men2024baseropeboundscontext}, and developed techniques to extend context length efficiently \citep{hooper2024kvquant10millioncontext,pmlr-v235-ding24i}. Others have focused on improving inference efficiency \citep{jiang2024minference10acceleratingprefilling,tang2024questqueryawaresparsityefficient} or enhancing long-context utilization and mitigating information loss \citep{zhang2024middlelanguagemodelsuse,lin2024mixtureincontextexpertsenhance,zhang2024chainagentslargelanguage}. Table \ref{tab:literature_review} highlights numerous benchmarks developed to evaluate long-context LLMs on different tasks. 

We introduce the \fullname{} (\frameworkshortnamebold{}) framework, outlined in Figure \ref{fig:our_pipeline}. Our approach uses knowledge-graph document representation to \emph{programmatically} extract question-answer pairs of varying complexity \emph{from gold annotations}. We define \emph{question templates} and determine their complexity through three dimensions: the number of hops, the use of set operations, and the plurality of the answer. This structured approach ensures scalability, flexibility, and controlled dataset construction, enabling a more precise evaluation of long-context LLMs.

We apply \frameworkshortname{} to 170 credit agreements, resulting in \datasetshortnamebold{}, a dataset of 20,139 QA pairs categorized by five complexity levels. We use it to evaluate 16 long-context LLMs and find that even on simple factual queries, some models fail to extract basic information; performance further drops with multi-hop paths and questions involving set operations. Our analysis shows that set-based comparisons are especially challenging, with high rates of ``Not Found'' responses and semantic misinterpretation.

Our contributions are as follows. (1) We present a framework based on knowledge graphs, enabling systematic extraction of QA pairs at varying complexity levels. (2) We release a portion of the constructed dataset\footnote{\url{https://huggingface.co/datasets/gtfintechlab/KG-MuLQA-D}} to encourage future work (see Appendix \ref{app:dataset_release} for details). (3) We release our inference pipeline and benchmarking codes\footnote{\url{https://github.com/gtfintechlab/KG-MuLQA}} to facilitate reproducible evaluation of long-context models.

\section{The \frameworkshortnamebold{} Benchmark}
\label{sec:our_benchmark}

\frameworkshortname{} systematically extracts structured QA pairs across varying complexity levels for long-context evaluation. Using credit agreements (Section \ref{subsec:credit_agreeements}) annotated with a structured label schema (Section \ref{subsec:annotations}, Appendix \ref{app:annotation_guide}), we construct document-level knowledge graphs (Section \ref{subsec:knowledge_graph_representation}) to enable systematic query construction. Questions are designed along three dimensions -- multi-hop reasoning, set operations, and answer plurality (Section \ref{subsec:question_complexity_dimensions}) -- ensuring scalable, diverse, and automated question extraction (Section \ref{subsec:qa_pairs_extraction}). The dataset balances complexity and document coverage (Appendix \ref{subapp:complexity_vs_coverage}), making it a robust long-context LLM benchmark.

\subsection{Credit Agreements}
\label{subsec:credit_agreeements}

We sampled 17 credit agreements per year from 2013 to 2022 from SEC EDGAR, totaling 170 documents. Table \ref{tab:annotation_stats} includes their length by token and sentence count. Their suitability for long-context LLM benchmarking is detailed in Appendix \ref{subapp:credit_agreement_suitability}.

\subsection{Annotations}
\label{subsec:annotations}

Our annotation schema captures key entities and relationships in credit agreements, defining seven label types covering persons, organizations, roles, and locations to represent their structure. We outline label definitions in Appendix \ref{subapp:labels}, placement rules and relationship guidelines in Appendix \ref{subapp:general_rules}, while Table \ref{tab:annotation_stats} includes the frequency statistics for each entity and relation type across documents. Each document was independently annotated by 3 annotators. Between the second and third (final) rounds of annotating, the inter-annotator agreement reached 80.76\%, indicating substantial consistency (see Appendix \ref{subapp:annotation_process_details} for annotation process details).

\subsection{Knowledge Graph Representation}
\label{subsec:knowledge_graph_representation}

\begin{figure*}[t]
    \centering
    \includegraphics[width=0.98\linewidth]{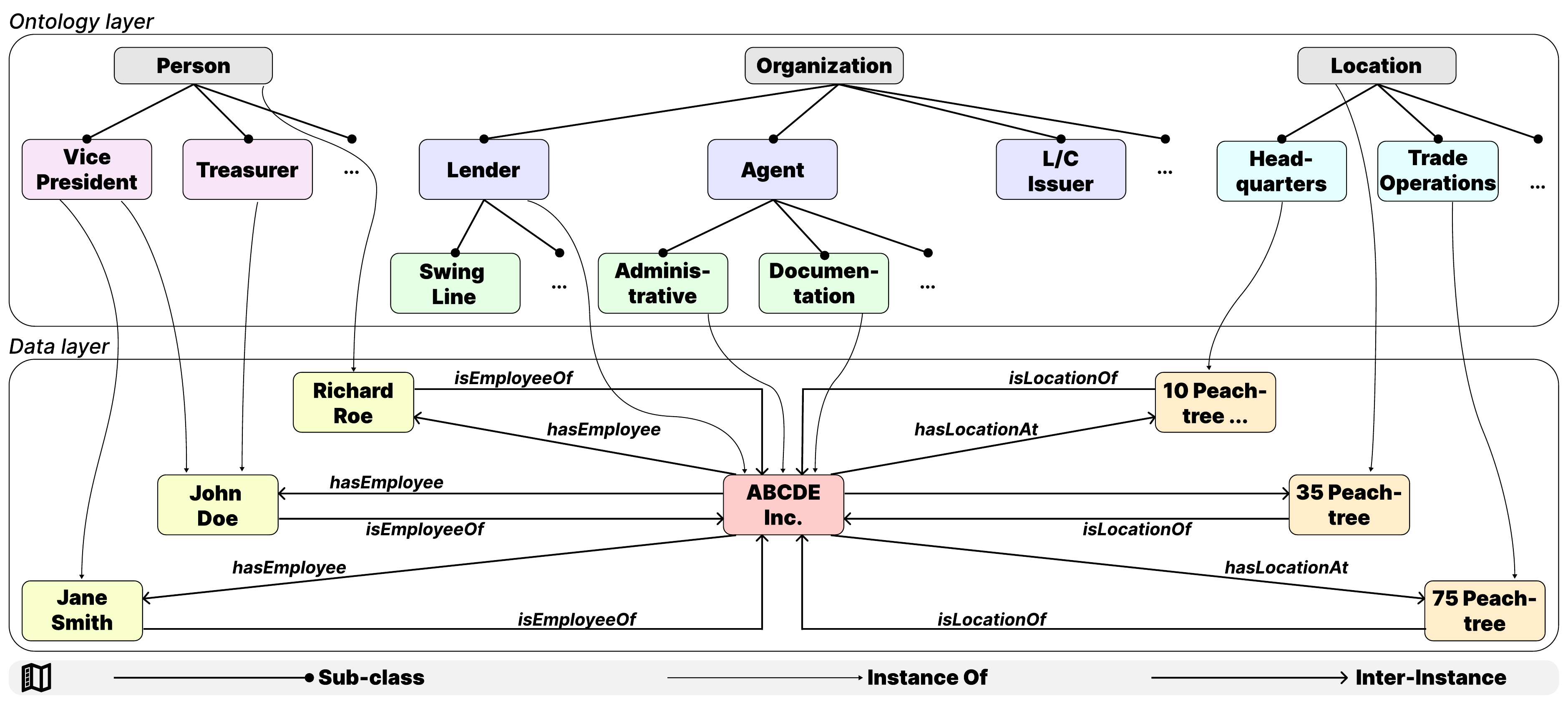}
    \caption{Knowledge graph representation of annotated credit agreements. The ontology layer defines high-level entity types and their sub-classes (e.g., Agent, Lender), while the data layer contains document-specific instances connected via labeled relations (e.g., hasEmployee, isLocationOf). This abstraction supports systematic QA extraction across documents by removing company-specific constraints. All entity names in this figure, including individuals, organizations, and addresses, are fictional and used for illustrative purposes only.}
    \label{fig:knowledge_graph}
\end{figure*}

To enable systematic and scalable QA extraction, we transform annotated credit agreements into knowledge graph representations. These graph encode entities of people, organizations, roles, and locations, along with their relationships, in a structured and queryable form. Figure \ref{fig:knowledge_graph} illustrates a graph which captures that \colorbox[rgb]{0.95,0.95,0.78}{\texttt{Jane Smith}} holds the position of \colorbox[rgb]{0.88,0.76,0.91}{\texttt{Vice President}} at \colorbox[rgb]{0.96,0.71,0.71}{\texttt{ABCDE Inc.}} -- an organization that serves as a \colorbox[rgb]{0.75,0.71,0.96}{\texttt{Lender}} as well as an \colorbox[rgb]{0.71,0.91,0.71}{\texttt{Administrative}} and \colorbox[rgb]{0.71,0.91,0.71}{\texttt{Documentation}} \colorbox[rgb]{0.75,0.71,0.96}{\texttt{Agent}}, and has its \colorbox[rgb]{0.72,0.90,0.97}{\texttt{Headquarters}} at \colorbox[rgb]{0.97,0.90,0.71}{\texttt{10 Peachtree...}}. By abstracting away document-specific phrasing while preserving these logical links, the graphs support consistent interpretation across agreements and form the basis for creating questions of varying complexity through structured queries.

\begin{table}[t]
    \caption{Statistics of the source documents and their structured annotations used to construct \datasetshortname{}. The table summarizes token and sentence counts, as well as the distribution of annotated entities and relations per document. The low minimum values are due to an edge case discussed in Appendix \ref{subapp:main_edge_cases}.}
    \label{tab:annotation_stats}
    \centering
    \resizebox{\linewidth}{!}{ 
    \begin{tabular}{lccc}
        \toprule
        \textbf{Stats per doc (total 170 docs)} & \textbf{Min} & \textbf{Avg} & \textbf{Max}\\
        \midrule
        \multicolumn{4}{l}{\textbf{Documents}} \\
        \midrule
        Tokens & 32,562 & 83,694 & 233,207 \\
        Sentences & 470 & 1,066 & 3,541 \\
        \midrule
        \multicolumn{4}{l}{\textbf{Entities}} \\
        \midrule
    \colorbox[rgb]{0.96,0.71,0.71}{\texttt{Org Name}}       & 2 & 27.52 & 390  \\ 
    \colorbox[rgb]{0.75,0.71,0.96}{\texttt{Org Role}}       & 1 & 13.82 & 52 \\
    \colorbox[rgb]{0.71,0.91,0.71}{\texttt{Org Sub-Role}}   & 0 & 4.26  & 35 \\
    \colorbox[rgb]{0.95,0.95,0.78}{\texttt{Person Name}}    & 0 & 12.55 & 88 \\
    \colorbox[rgb]{0.88,0.76,0.91}{\texttt{Person Position}}& 0 & 13.17 & 88 \\
    \colorbox[rgb]{0.97,0.90,0.71}{\texttt{Location}}       & 0 & 5.61  & 168 \\
    \colorbox[rgb]{0.72,0.90,0.97}{\texttt{Location Type}}  & 0 & 0.45  & 7 \\
        \midrule
        \multicolumn{4}{l}{\textbf{Relations}} \\
        \midrule
    \colorbox[rgb]{0.96,0.71,0.71}{\texttt{Org Name}} \( \rightarrow \) \colorbox[rgb]{0.75,0.71,0.96}{\texttt{Org Role}}         & 1 & 21.17 & 79 \\
    \colorbox[rgb]{0.75,0.71,0.96}{\texttt{Org Role}} \( \rightarrow \) \colorbox[rgb]{0.71,0.91,0.71}{\texttt{Org Sub-Role}} & 0 & 3.74  & 32 \\
    \colorbox[rgb]{0.96,0.71,0.71}{\texttt{Org Name}} \( \rightarrow \) \colorbox[rgb]{0.95,0.95,0.78}{\texttt{Person Name}}     & 2 & 12.55 & 88 \\
    \colorbox[rgb]{0.96,0.71,0.71}{\texttt{Org Name}} \( \rightarrow \) \colorbox[rgb]{0.88,0.76,0.91}{\texttt{Person Position}} & 0 & 13.17 & 88 \\
    \colorbox[rgb]{0.95,0.95,0.78}{\texttt{Person Name}} \( \rightarrow \) \colorbox[rgb]{0.88,0.76,0.91}{\texttt{Person Position}} & 0 & 12.93 & 86 \\
    \colorbox[rgb]{0.96,0.71,0.71}{\texttt{Org Name}} \( \rightarrow \) \colorbox[rgb]{0.97,0.90,0.71}{\texttt{Location}}        & 0 & 6.32  & 178 \\
    \colorbox[rgb]{0.97,0.90,0.71}{\texttt{Location}} \( \rightarrow \) \colorbox[rgb]{0.72,0.90,0.97}{\texttt{Location Type}}   & 0 & 0.45  & 18 \\
        \bottomrule
    \end{tabular}
    }
\end{table}

We construct knowledge graphs using the Resource Description Framework\footnote{\url{https://www.w3.org/TR/rdf11-concepts/}} (RDF), which represents data as subject-predicate-object triples, forming a directed labeled graph for knowledge representation. Stored in Turtle (*.ttl) format for readability, this structure ensures scalability and efficient querying across documents using SPARQL\footnote{\url{https://www.w3.org/TR/sparql11-query/}} (see Appendix \ref{app:knowledge_graph_construction} for KG construction details).

\subsection{Question Templates and Complexity Dimensions}
\label{subsec:question_complexity_dimensions}

\begin{table*}[t]
    \caption{This table illustrates the question templates used to construct \datasetshortname, structured along three dimensions: plurality (P), number of hops (H), and set operations (\#SO). It includes example templates, corresponding knowledge graph query paths, and logical operations involved. These dimensions are used to compute the overall complexity level for each QA pair. The full list of templates can be found in Appendix \ref{app:all_templates}.}
    \label{tab:sample_templates}
    
    \renewcommand{\arraystretch}{1.2}  
    \setlength{\abovecaptionskip}{6pt}  
    \setlength{\belowcaptionskip}{10pt} 
    \scriptsize
    \setlength{\tabcolsep}{4pt}
    \begin{center}
    \vspace{0.5em}  
    \begin{tabular}{|c|c|c|p{0.53\textwidth}|p{0.15\textwidth}|p{0.13\textwidth}|}
        \hline
        \rowcolor{gray!15}  
        \textbf{P} & \textbf{H} & \textbf{\#SO} & \textbf{Example Template} & \textbf{Hop Path} & \textbf{Set Operation} \\
        \hline
        0 & 1 & 0 & What is the position of [Person Name]? & Person→Position & None \\
        \hline
        1 & 1 & 0 & What are the positions of [Person Name]? & Person→Position & None \\
        \hline
        0 & 1 & 1 & What position is held by both [Person A] and [Person B]? & Person→Position & $A \cap B$ \\
        \hline
        0 & 1 & 2 & What position does [Person A] hold that [Person B] doesn't? & Person→Position & $A \cap \neg B$ \\
        \hline
        0 & 1 & 3 & What position is held by [Person A] but not [Person B] or [Person C]? & Person→Position & $A \cap \neg B \cap \neg C$   \\
        \hline
        1 & 1 & 1 & What are the positions held by both [Person A] and [Person B]? & Person→Position & $A \cap B$ \\
        \hline
        1 & 1 & 2 & What are the positions held by [Person A] but not [Person B]? & Person→Position & $A \cap \neg B$ \\
        \hline
        1 & 1 & 3 & What are the positions held by [Person A] but not [Person B] or [Person C]? & Person→Position &  $A \cap \neg B \cap \neg C$ \\
        \hline
        0 & 2 & 0 & Who is the [Position] of [Org]? & Org→Pos→Per & None \\
        \hline
        1 & 2 & 0 & What are the roles of [Org] in the agreement where [Person] is employed? & Per→Org→Role & None \\
        \hline
        0 & 2 & 1 & Who is the [Position A] and [Position B] of [Org]? & Org→Pos→Per & $A \cap B$ \\
        \hline
        1 & 2 & 1 & Who are both [Position A]s and [Position B]s of [Org]? & Org→Pos→Per & $A \cap B$ \\
        \hline
        0 & 3 & 0 & Who is the [Position] of the company associated with [Location]? & Loc→Or→Ps→Pe & None \\
        \hline
        1 & 3 & 0 & Who are the [Position]s of the company associated with [Location]? & Loc→Or→Ps→Pe & None \\
        \hline
        0 & 3 & 1 & Who is both [Position A] and [Position B] of the company associated with [Location]? & Loc→Or→Ps→Pe & $A \cap B$  \\
        \hline
        0 & 3 & 2 & Who is [Position A] but not [Position B] of the company associated with [Location]? & Loc→Or→Ps→Pe & $A \cap \neg B$ \\
        \hline
        0 & 3 & 3 & Who is [Position A] and [Position B] but not [Position C] of the company associated with [Location]? & Loc→Or→Ps→Pe & $A \cap  B \cap \neg C$ \\
        \hline
        1 & 3 & 1 & Who are both [Position A] and [Position B] of the company associated with [Location]? & Loc→Or→Ps→Pe & $A \cap B$ \\
        \hline
        1 & 3 & 2 & Who are [Position A]s but not [Position B]s of the company associated with [Location]? & Loc→Or→Ps→Pe & $A \cap \neg B$ \\
        \hline
    \end{tabular}
    \end{center}
\end{table*}

We begin by identifying all possible 1-hop QA templates from the knowledge graph, illustrated in Figure \ref{fig:knowledge_graph}: questions formed from direct relationships between two entities, such as ``\textit{What is the position of} [\colorbox[rgb]{0.95,0.95,0.78}{\texttt{Person Name}}]?" or ``\textit{Who is the representative of} [\colorbox[rgb]{0.96,0.71,0.71}{\texttt{Org Name}}]\textit{?}". These base templates are constrained to yield singular answers, forming the simplest form of retrieval. It should be noted that not all templates are universally applicable -- for instance, the latter question is only valid when the document specifies exactly one representative for the given organization. We then systematically expand these templates along three dimensions.

\paragraph{Number of Hops} Increasing retrieval depth via intermediate nodes, e.g., ``\textit{Who is the} [\colorbox[rgb]{0.88,0.76,0.91}{\texttt{Person Position}}] \textit{of the organization located at} [\colorbox[rgb]{0.97,0.90,0.71}{\texttt{Location}}]\textit{?}".

\paragraph{Set Operations} Introducing comparisons across entities using set logic, e.g., ``\textit{What are the positions held by} [\colorbox[rgb]{0.95,0.95,0.78}{\texttt{Person Name A}}] \textit{but not} [\colorbox[rgb]{0.95,0.95,0.78}{\texttt{Person Name B}}] \textit{or} [\colorbox[rgb]{0.95,0.95,0.78}{\texttt{Person Name C}}]\textit{?}".

\paragraph{Plurality}  Moving from expecting a singular answer to allowing multiple valid answers, e.g., ``\textit{What companies are the} [(\colorbox[rgb]{0.71,0.91,0.71}{\texttt{Org Sub-Role}} +) \colorbox[rgb]{0.75,0.71,0.96}{\texttt{Org Role}}] \textit{in the agreement?}".

\subsection{Extraction of QA Pairs from Templates}
\label{subsec:qa_pairs_extraction}

Table \ref{tab:sample_templates} illustrates how questions are systematically constructed by incrementally increasing complexity across three dimensions introduced in Section \ref{subsec:question_complexity_dimensions}: number of hops (H), plurality (P), and number of set operations (\#SO). Each row in the table corresponds to a specific question template, with associated values for P, H, and \#SO, the query path in a KG, and the set operation logic used.

Question instances are constructed by dynamically substituting entities into these templates using SPARQL queries over the knowledge graph described in Section \ref{subsec:knowledge_graph_representation}. This ensures precise and scalable QA pair creation, \textbf{\textit{with answer labels grounded in annotated relationships}}. While our framework supports arbitrarily complex combinations, we observed that highly compositional templates often become inapplicable to real documents. For instance, there are no individuals in an agreement who hold three distinct positions or no organization with the combination of roles required to fulfill a complex query. Thus, we have to limit template depth since such complex conditions are typically not present in the source documents. The list of all the created templates can be found in Appendix \ref{app:all_templates}.

To support evaluation, we define a composite question complexity level \( L \! = \! H\ \! +\! P\! +\! \#SO\), and group QA pairs into three categories. \textbf{Easy} (\( L \! = \! 1\)): simple, single-hop, singular-answer queries. \textbf{Medium} (\( 2 \! \leqslant \! L \! \leqslant \! 4 \)): questions with moderate complexity via hops, set operations, or plurality. \textbf{Hard} (\( L \! = \! 5 \)): questions with the most complex combinations of all three dimensions.

The resulting dataset, \datasetshortname, comprises 20,139 QA pairs from 170 credit agreements, surpassing most long-context QA benchmarks in size (Table \ref{tab:literature_review}). This scale is achieved through efficient question extraction from structured annotations using reusable templates.

\section{Evaluation}
\label{sec:evaluation}

In this section, we describe the setup for our experiments. We evaluate multiple long-context LLMs listed in Section \ref{subsec:baselines} across different question complexity levels, as defined in Section \ref{subsec:question_complexity_dimensions}. Performance is measured using four metrics detailed in Section \ref{subapp:metrics}. In addition, we conduct human evaluation of LLM responses on a small sample of QA pairs (Section \ref{subapp:human_evaluation_metrics}) to discover that F1-score and the LLM-as-a-Judge metrics correlate with human's expectations most. Our prompting strategy is outlined in Section \ref{subapp:prompting_strategy}. The results of our level-based evaluation are presented in Table \ref{tab:main_results}, with the analysis provided in Section \ref{sec:analysis}.

\subsection{Baselines}
\label{subsec:baselines}

We evaluate 16 proprietary and open-weight LLMs on \datasetshortname{} benchmark.

\paragraph{Proprietary LLMs:} Gemini-2.0-Flash \& Gemini-2.0-Flash-Lite \citep{team2024gemini}, Gemini-1.5-Pro \& Gemini-1.5-Flash \citep{team2023gemini, team2024gemini}, and GPT-4o \& GPT-4o-Mini \citep{achiam2023gpt, hurst2024gpt}. 

\paragraph{Open-Weight LLMs:} DeepSeek-R1 \citep{deepseekai2025deepseekr1incentivizingreasoningcapability}, DeepSeek-V3 \citep{liu2024deepseek}, GPT OSS 120B \citep{openai2025gptoss120bgptoss20bmodel}, Llama-4-Maverick-17B-128E-Instruct-FP8 \citep{meta2024llama4maverickfp8}, Llama-4-Scout-17B-16E-Instruct \citep{meta2024llama4scout}, Llama-3.1-405B-Instruct-Turbo \& Llama-3.1-70B-Instruct \& Llama-3.1-8B-Instruct \citep{dubey2024llama}, Qwen3-235B-Instruct \citep{yang2025qwen3technicalreport}, Qwen2-72B-Instruct \citep{yang2024qwen2}. 

Given that our documents average 80k tokens, we restrict our evaluation to models supporting a context length of at least 128k tokens. Nonetheless, even long-context models require a carefully designed prompting strategy to extract relevant answers effectively (see Section \ref{subapp:prompting_strategy}). We made several attempts to evaluate Gemma-2-8k \citep{team2024gemma}, Mixtral-Small-32k, and Mixtral-8×22-65k \citep{jiang2024mixtral}, but these models returned ``\texttt{Not found}'' in the vast majority of the cases, making systematic evaluation infeasible.

For proprietary models and those accessed via Together.AI\footnote{\url{https://www.together.ai/}}, we utilized the \texttt{LangChain}\footnote{\url{https://www.langchain.com/}} framework, while open-weight models were employed using \texttt{vLLM} \citep{kwon2023efficient} for efficient inference. We set the temperature to 0.0 to support reproducibility of our results.

\subsection{Experiment Setup}
\label{subsec:experiment_setup}

\begin{figure*}[t]
    \centering
    \includegraphics[width=\linewidth]{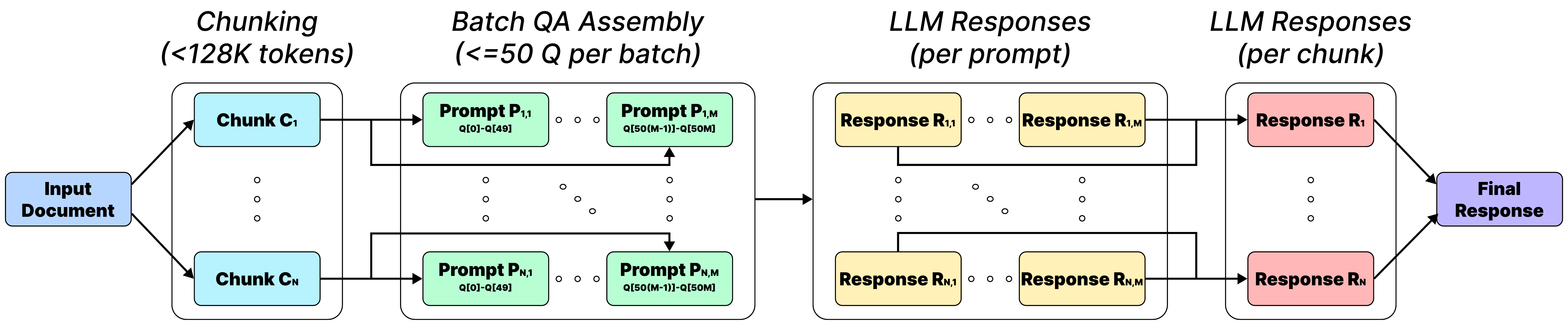}
    \caption{This figure illustrates the multi-stage process used to evaluate long-context LLMs: documents are split into chunks (if exceeding 128K tokens), paired with question batches (\( \leqslant \)50), and sequentially fed to the model. Model responses are collected per chunk and per question batch to ensure evaluation scalability across large documents.}
    \label{fig:prompting_strategy}
\end{figure*}

Our evaluation setup is designed to handle long financial documents and large batches of questions in a way that fits the context length constraints of modern language models. As seen in Figure \ref{fig:prompting_strategy}, we split a document into chunks if it exceeds 128K tokens: most of the evaluated models have context length of 128K tokens, so \textbf{it is impossible to evaluate these models without chunking}. For the larger models, we \textbf{have to} keep the same chunking strategy to ensure a fair comparison between models (see Appendix \ref{subapp:chunk_based_evaluation} for the effect of chunking on evaluation). Then, we prompt the model with the document (or chunk) and a batch of 50 questions associated with it: long-context evaluation is very costly, so batching allows at most 3 inference runs per document (given almost 150 questions for some documents). This design ensures consistent and scalable evaluation across documents of varying lengths (see Appendix \ref{subapp:prompting_strategy} for further details).

We use four evaluation metrics: F1 score, Edit Distance, Cosine Similarity, and LLM-as-a-Judge. Among these, F1 and LLM-as-a-Judge are highlighted in Table \ref{tab:main_results} for their complementary strengths: F1 quantifies exact token-level overlap, while LLM-J captures semantic correctness with human-aligned scoring (see Appendix \ref{subapp:metrics} for details). Both metrics show the highest correlation with human evaluation metrics (see Appendix \ref{subapp:human_evaluation_metrics}).

\subsection{Main Results}
\label{subsec:main_results}

As seen in Table \ref{tab:main_results}, the results for simple questions are straightforward to interpret: more advanced models mostly get better scores. For medium questions, DeepSeek-V3 outperforming DeepSeek-R1 and Gemini-1.5-Pro outperforming Gemini-2.0 mainly comes from fewer abstentions and ``\texttt{Not Found}'' cases, which raises recall and F1 but doesn’t reflect stronger reasoning. Llama-3.1 405B behaves similarly: it answers more often, boosting both metrics through coverage rather than logic. In contrast, Qwen2’s high judge score but low F1 reflects a different pattern: it tends to produce verbose, loosely correct text that seems semantically reasonable, so the judge rates it highly even when exact matches are wrong or incomplete. The GPT OSS 120B, however, gets good performance on medium difficulty from its ability to extract entities while minimizing errors, as it strikes a balance between answering and abstaining.

For hard questions on proprietary models, Gemini-2.0-Flash shows high judged quality but lower F1, while Flash-Lite reverses this trend; Gemini-1.5-Pro remains balanced. This difference comes from how often each model answers: more attempts raise F1 through partial overlaps with the gold, while fewer but cleaner attempts earn better judged quality (given responses are not verbose).

For open-weight models, Llama-4-Maverick achieves a high LLM-J score with moderate F1 for answering selectively with short, factual fragments that look semantically right but miss full token matches. Llama-3.1 405B, by contrast, produces longer and more complete answers that align better with references, yielding higher F1 while maintaining strong judged coherence, hence demonstrating the best performance with hard questions.

Overall, as task complexity increases, models split between answering broadly and answering cautiously. This divergence weakens the correlation between F1 and judge scores, since literal overlap and perceived semantic adequacy start reflecting different behaviors.

\section{Analysis}
\label{sec:analysis}

\begin{table*}[t]
    \caption{This table presents the performance of 16 LLMs, evaluated across Easy, Medium, and Hard question categories. The metrics include the F1 Score and the LLM-as-a-Judge rating, capturing both token-level accuracy and semantic correctness (see the appendix \ref{subapp:metrics} for extended evaluation). The results reveal a consistent decline in performance as question complexity increases, with notable model-specific strengths and weaknesses. $^{\ast}$ denotes the models evaluated on a smaller subset due to cost constraints (see Appendix \ref{subapp:cost_constrained_models} for details).}
    \label{tab:main_results}
    
    \centering
    \resizebox{\linewidth}{!}{ 
    \rowcolors{4}{gray!20}{white} 
    \begin{tabular}{lcc|cc|cc|cc}
        \hline
        & & & \multicolumn{2}{|c}{\textbf{Easy}} & \multicolumn{2}{|c}{\textbf{Medium}} & \multicolumn{2}{|c}{\textbf{Hard}} \\
        \cline{4-9}
        \textbf{Model} & \textbf{Size} & \textbf{Context} & F1 & LLM-as- & F1 & LLM-as- & F1 & LLM-as- \\
        & & \textbf{Length} & Score $\uparrow$ & a-Judge $\uparrow$ & Score $\uparrow$ & a-Judge $\uparrow$ & Score $\uparrow$ & a-Judge $\uparrow$ \\
        \hline
        
        \multicolumn{9}{l}{\textbf{Proprietary LLMs}} \\
        Gemini-2.0-Flash$^{\ast}$ & -- & 1M & \textbf{0.6748} & \textbf{4.0000} & 0.3294 & 2.4820 & 0.1172 & \textbf{2.2838} \\ 
        Gemini-2.0-Flash-Lite$^{\ast}$ & -- & 1M & \underline{0.5908} & \underline{3.4172} & 0.3883 & \underline{2.6497} & \textbf{0.1943} & 1.5135 \\
        Gemini-1.5-Pro$^{\ast}$ & -- & 2M & 0.5410 & 3.2448 & \textbf{0.4577} & \textbf{2.8436} & \underline{0.1791} & \underline{1.6986} \\
        Gemini-1.5-Flash$^{\ast}$ & -- & 1M & 0.5103 & 3.1207 & \underline{0.3892} & 2.5130 & 0.1400 & 1.5411 \\
        GPT-4o$^{\ast}$ & -- & 128K & 0.5272 & 3.2034 & 0.3783 & 2.5451 & 0.1178 & 1.4795 \\
        GPT-4o-Mini$^{\ast}$ & -- & 128K & 0.4723 & 2.9690 & 0.2732 & 2.0681 & 0.0274 & 1.1370 \\
        
        \hline
        
        \multicolumn{9}{l}{\textbf{Open-weight LLMs}} \\
        DeepSeek-R1$^{\ast}$ & 685B & 128K & \textbf{0.6269} & \textbf{3.6965} & 0.3293 & 2.6354 & \underline{0.2609} & 2.0736 \\
        DeepSeek-V3 & 671B & 128K & \underline{0.6218} & 3.5668 & \textbf{0.4796} & \underline{3.0210} & 0.0856 & 1.4795 \\
        GPT OSS & 120B & 128K & 0.5885 & \underline{3.5839} & \underline{0.4684} & \textbf{3.1661} & 0.1136 & 1.4888 \\
        Llama-4-Maverick-Instruct$^{\ast}$ & 400B & 1M & 0.5552 & 3.4482 & 0.3848 & 2.6976 & 0.2031 & \textbf{3.1351} \\
        Llama-4-Scout-Instruct$^{\ast}$ & 109B & 1M & 0.4993 & 3.2138 & 0.2215 & 1.8772 & 0.0472 & 1.5270 \\
        Llama-3.1-Instruct-Turbo$^{\ast}$ & 405B & 128K & 0.5472 & 3.4793 & 0.4204 & 2.6377 & \textbf{0.3651} & \underline{2.7297} \\
        Llama-3.1-Instruct & 70B & 128K & 0.4955 & 3.1373 & 0.4017 & 2.6722 & 0.1639 & 2.3425 \\
        Llama-3.1-Instruct & 8B & 128K & 0.3891 & 2.6996 & 0.2962 & 2.2332 & 0.1169 & 1.4315 \\
        Qwen3-Instruct & 235B & 262K & 0.5243 & 3.2693 & 0.2760 & 2.4460 & 0.1240 & 1.9578 \\
        Qwen2-Instruct & 72B & 128K & 0.3886 & 3.3884 & 0.3100 & 2.9078 & 0.0762 & 1.3288 \\
        \hline
    \end{tabular}
    }
\end{table*}

\subsection{Overall Error Trends}
\label{subsec:error_trends}

\begin{figure}[t]
    \centering
    \includegraphics[width=\linewidth]{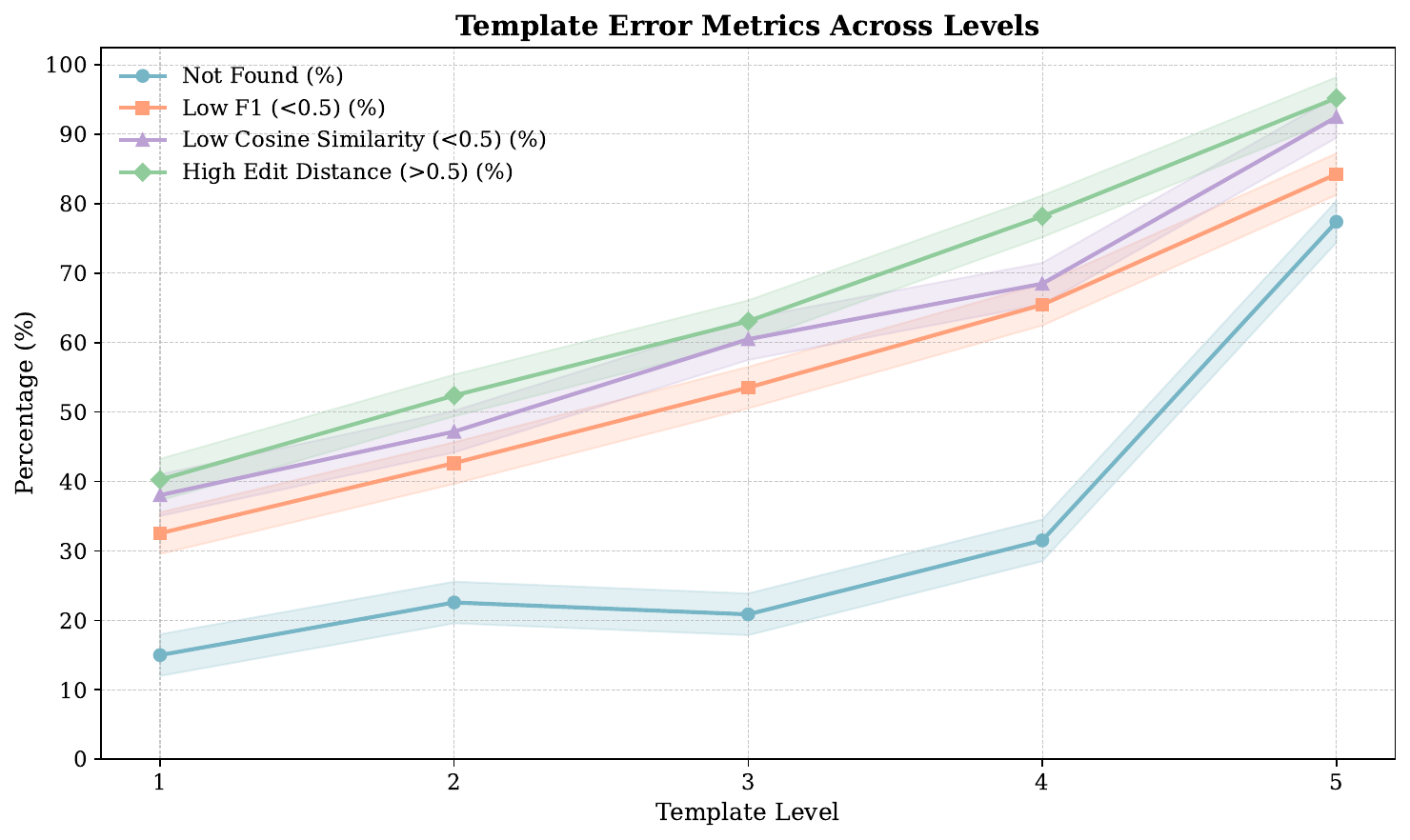}
    \caption{Trends in error types across complexity levels. The template level is given by \(P + H + \#SO\).}
    \label{fig:error_metrics_comparison}
\end{figure}

As question complexity increases, the LLM's ability to retrieve and generate correct responses degrades markedly. With increase in number of hops or set operations, the percentage distribution of each error type increases (see Figure \ref{fig:error_metrics_comparison} for details). 

\begin{itemize}[leftmargin=*,nosep]
     \item\textbf{Not Found Responses.} Increases from 15.00\% at Level 1 to 77.39\%, reflecting increasing retrieval difficulty.
     \item\textbf{Low F1 Scores.} Rises from 32.55\% to 84.25\%, indicating less accurate extraction.
     \item\textbf{Low Cosine Similarity.} Jumps from 38.03\% to 92.47\%, showing semantic drift.
     \item\textbf{High Edit Distance.} Grows from 40.26\% to 95.21\%, reflecting surface mismatch.
\end{itemize}

\subsection{Error Analysis}
\label{subsec:error_analysis}

We categorize observed LLM failures into four major types, each of which presents recurring challenges as question complexity increases. For detailed error analysis on examples from each category, see Appendix~\ref{app:error_analysis}, Table~\ref{tab:error_analysis_examples}.

\noindent \textbf{Misinterpretation of Semantics}
LLMs frequently misinterpret task-specific terms such as ``location type'' or ``position''. Rather than grounding responses in document context, models often default to general world knowledge (e.g., answering ``city'' for a location type), indicating weak comprehension of domain-specific semantics.

\noindent \textbf{Implicit Information Gaps}
The model struggles to extract answers that are not explicitly stated but implied through document structure, such as signature blocks or formatting cues. For instance, it fails to associate a person with an organization unless the relationship is stated verbatim, ignoring visual or positional indicators that a human reader would readily infer. We consider it a potential reason for why models perform slightly better on full documents than in the oracle setting for hard questions (see Appendix \ref{subapp:rag_oracle_results} for details).

\noindent \textbf{Set Operation Failures} To examine the impact of the number of set operations on model performance, we conducted an ablation experiment (Figure \ref{fig:set_ops_plot}) by varying one dimension within our three-dimensional parameter space while holding the others constant. Specifically, we increased the number of set operations while keeping the plurality and hop dimensions fixed. We observe that an increase in the number of set operations consistently resulted in a decline in performance metrics. Questions involving comparisons, intersections, or exclusions across entities consistently yield low F1 scores. The model fails to isolate shared or distinct roles when multiple organizations or individuals are involved, indicating limited capacity for multi-hop or set-based reasoning.

\noindent \textbf{Long-Context Retrieval Errors} Despite the presence of answers in the document, models often return ``\texttt{Not Found}'', especially when relevant information appears in the middle of the context. This aligns with the ``lost-in-the-middle'' phenomenon \citep{liu-etal-2024-lost}, where models disproportionately focus on the beginning or end of long texts.

Our error analysis reveals that while LLMs perform adequately on simple, single-step queries, they falter on multi-hop, set-operation, and implicit-knowledge tasks -- particularly those requiring retrieval of details buried in the ``middle" of the context. Addressing these shortcomings will likely involve a combination of targeted fine-tuning, improved context chunking or retrieval modules, and stronger in-context reasoning prompts.

\section{Related Work}
\label{sec:related_work}

Answering questions about a long document requires, first of all, \textbf{\textit{retrieval}} of relevant information piece(-s). Based on the nature of the questions asked, answering them might require \textbf{\textit{reasoning}} over the retrieved piece(-s), while reasoning itself can be defined differently across benchmarks. For example, CLongEval \citep{qiu-etal-2024-clongeval},  \( \infty \)Bench \citep{zhang-etal-2024-bench}, MMLongBench-DOC \citep{ma2024mmlongbenchdocbenchmarkinglongcontextdocument} and Michelangelo \citep{vodrahalli2024michelangelolongcontextevaluations} treat long-context reasoning as synthesizing a conclusion from lengthy inputs, and the former one evaluates it via two task types: abstraction (generate content not explicitly in the source) and extraction (identify and copy from the input). MARATHON \citep{zhang-etal-2024-marathon} frames long-context reasoning as choosing a single correct option from long contexts in a multiple-choice setup with human-verified, misleading distractors. DocFinQA \citep{reddy-etal-2024-docfinqa} and DocMath-Eval \citep{zhao-etal-2024-docmath} evaluate long-context numerical reasoning: models find the relevant evidence and carry out multi-step numerical and logical reasoning over text and tables to produce the answer. In our study, we focus on \textbf{\textit{multi-hop reasoning}}: combining multiple evidence pieces by following multi-step relations between entities to reach an answer. It allows to directly measure a model’s ability to track dependencies, integrate facts, and maintain coherence across distant spans: abilities that are central to robust long-context comprehension.

Prior long-context benchmarks typically rely on manual QA construction, which is very costly and time-consuming. RULER \citep{hsieh2024rulerwhatsrealcontext}, mLongRR \citep{agrawal-etal-2024-evaluating}, and $\infty$Bench \citep{zhang-etal-2024-bench} all evaluate retrieval by inserting target facts/keys into long distractor contexts and probing whether models can recover or aggregate them at scale. M4LE \citep{kwan-etal-2024-m4le}, Michelangelo \citep{vodrahalli2024michelangelolongcontextevaluations}, and BABILong \citep{kuratov2024babilongtestinglimitsllms} programmatically synthesize evaluation items, using templates/simulations or latent-structure queries to automatically create large numbers of QA/tasks without manual question writing. We \textbf{\textit{deterministically extract}} multi-level QA directly from knowledge-graph representations of annotated documents, varying hops, set operations, and plurality via structured queries. We do not insert synthetic facts, preserving real document distribution, and we do not rely on LLM-written items that would require human verification. MuSiQue \citep{trivedi-etal-2022-musique} also follows a systematic pipeline, but it first requires manually creating the base single-hop questions and then composing them; to build a new set, those easy questions must be recreated, whereas our pipeline is fully automated from document annotations.

Most existing benchmarks do not define explicit question complexity levels, treating all QA instances as uniformly difficult. Several recent datasets \citep{gupta-etal-2024-systematic,zhao-etal-2024-docmath,zhu-etal-2024-fanoutqa,hsieh2024rulerwhatsrealcontext,kuratov2024babilongtestinglimitsllms} vary task difficulty by changing task type or length, distributing supporting facts in long contexts, or altering context length and needle placement. We introduce a \textbf{\textit{granular notion of question complexity}}, defined along multiple dimensions (multi-hop reasoning, set operations, and answer plurality), allowing systematic control over difficulty and enabling fine-grained error analysis (Section \ref{subsec:error_analysis}) that reveals distinct model failure modes. While MuSiQue also organizes questions by reasoning depth (2–4 hops), it lacks the multi-dimensional decomposition we propose. Our formulation extends beyond hop count to capture qualitatively different reasoning types, which provides a richer understanding of model strengths and weaknesses.

\section{Discussion \& Conclusion}
\label{sec:discussion}

Our work introduces \frameworkshortname{}, a framework that programmatically constructs long-context QA benchmarks using knowledge-graph representations of real-world documents. We begin by annotating credit agreements with a structured schema to capture entities and relations, convert them into RDF-based knowledge graphs, and generate multi-level QA pairs via deterministic SPARQL templates that vary in hop count, set operations, and answer plurality. We then evaluate 16 long-context LLMs across these controlled complexity levels using 4 complementary metrics and perform human evaluation to validate metric alignment. Our quantitative and qualitative analyses reveal consistent weaknesses in multi-hop retrieval, set-based reasoning, and implicit relation grounding, establishing \frameworkshortname{} as a scalable and interpretable framework for assessing long-context comprehension.

\paragraph{Human Performance Evaluation} Appendix \ref{subapp:human_performance_eval} shows human performance remains stable across difficulty. This baseline both grounds our metric choices and indicates model errors stem from reasoning/retrieval limits rather than dataset artifacts.

\paragraph{Applicability and Generalizability} As discussed in Appendix \ref{subapp:credit_agreement_suitability}, the ability to process long documents is essential for real-world QA tasks. Our benchmark illustrates this challenge using credit agreements, where information is densely structured, scattered, and cross-referenced. Legal contracts, policy documents, patent filings, and other non-legal such as medical records or scientific articles all exhibit similarly complex entity structures, making our methodology broadly applicable. To illustrate the applicability and generalizability of our framework, we performed a full-scale ablation study (see Appendix \ref{app:ablation_study_medical}) on medical documents, including defining the annotation schema, constructing question templates, etc.

\section*{Limitations}

The created QA pairs are based on financial credit agreements only. As with most existing QA benchmarks, the benchmark favors deductive logic because the dataset includes graph-representable questions.

\section*{Ethical Considerations}
\label{sec:ethical_considerations}

We generally avoid mentioning specific agreements and their details. Instead, we use placeholders such as \texttt{ABCDE Inc.} for organization names, \texttt{Richard Roe} for person names, and \texttt{10 Peachtree...} for locations to illustrate the structure of agreement entities and   relations. However, in our annotation guide (Appendix \ref{app:annotation_guide}) and error analysis (Appendix \ref{app:error_analysis}), we include screenshots of real document annotations and mention specific examples to highlight the complexity of these agreements. Moreover, we store the raw HTML files of the agreements in our GitHub repository, enabling other researchers to reproduce our benchmark or conduct further experiments. As these documents are publicly available through the SEC’s EDGAR system, their inclusion for both illustrative and experimental purposes raises no ethical concerns.

\bibliography{custom}

\appendix

\section{Annotation Guide}
\label{app:annotation_guide}

\subsection{Suitability of Credit Agreements}
\label{subapp:credit_agreement_suitability}

Credit agreements are particularly well-suited for long-context evaluation because they are \textbf{\textit{rich in entities and relations}}, involving numerous organizations, individuals, roles, and locations that appear in varied forms throughout the document, making both retrieval and reasoning challenging. They also exhibit \textbf{\textit{ambiguous co-reference}}, as entities may be referred to inconsistently (e.g., ``Bank of America, N.A.'' vs. ``Bank of America''), demanding precise resolution of abbreviations and partial mentions. While \textbf{\textit{structured but not standardized}}, these documents follow broad templates (e.g., loan terms, covenants, defaults) but vary greatly in phrasing and layout, preventing reliance on fixed patterns and requiring genuine comprehension. Finally, their \textbf{\textit{real-world significance}} as legally binding records of multi-million-dollar transactions makes accurate understanding essential for assessing risk, ensuring compliance, and supporting financial and regulatory decisions. Supporting examples illustrating these aspects are provided in the latter subsections of this Appendix.

\subsection{Labels}
\label{subapp:labels}

Our annotation framework captures key entities and relationships within credit agreements, focusing on seven labels that define organizational structure, roles, and locations.

\paragraph{\colorbox[rgb]{0.96,0.71,0.71}{\texttt{Organization Name}}} The name of a company or financial institution mentioned in the agreement. \textit{Examples: ``Goldman Sachs'', ``JP Morgan Chase''.} 

\paragraph{\colorbox[rgb]{0.75,0.71,0.96}{\texttt{Organization Role}}} The role that an organization holds in the agreement. \textit{Examples: ``Lender'', ``Borrower'', ``Guarantor''.}

\paragraph{\colorbox[rgb]{0.71,0.91,0.71}{\texttt{Organization Sub-Role}}} A more specific classification within an Organization Role that provides additional detail. \textit{Examples: ``Lead Lender'', ``Administrative Agent''.}

\paragraph{\colorbox[rgb]{0.95,0.95,0.78}{\texttt{Person Name}}} The name of an individual explicitly mentioned in the agreement. \textit{Examples: ``John Doe'', ``Jane Smith''.}

\paragraph{\colorbox[rgb]{0.88,0.76,0.91}{\texttt{Person Position}}} The title or role held by a person within an organization. \textit{Examples: ``Vice President'', ``Loan Officer''.}

\paragraph{\colorbox[rgb]{0.97,0.90,0.71}{\texttt{Location}}} A physical address associated with an organization. \textit{Examples: ``35 Peachtree St, Atlanta, GA''.}

\paragraph{\colorbox[rgb]{0.72,0.90,0.97}{\texttt{Location Type}}} A classification of a Location, providing additional information on its function. \textit{Example: ``Headquarters'', ``Branch Office''.}

\subsection{General rules}
\label{subapp:general_rules}

\paragraph{Unique Entity Annotation} Organization Names, Person Names, and Locations are annotated only \textit{once per document}, even if they appear multiple times. Organization Roles, Organization Sub-Roles, Person Positions, and Location Types are annotated \textit{once for each unique entity} they describe.

\textit{Example}: In Figure \ref{fig:unique_entity_annotation}, ``BANK OF AMERICA, N.A.'' appears multiple times but is annotated only once. However, its roles, ``Administrative Agent'' and ``Lender'', are each annotated where they are explicitly stated. If another organization is a lender and/or an administrative agent, those role labels will be annotated as close as possible to the name label. Similarly for ``Alexandra M. Knights'' and her position ``Authorized Signatory''.

\paragraph{Ordered Search for Annotations} To maintain consistency, we follow a structured order when searching for annotations.

\begin{itemize}[leftmargin=*,nosep]

    \item \textit{Signature Box}: The signature box is the primary source for annotations, as it usually contains the most  organizations (e.g., Parent Borrowers, Lead Lenders) and their roles. In most cases, organizations are explicitly labeled with their roles within this section, making it the most reliable source.
    
    \item \textit{Headers}: If an entity is missing in the signature box, we check the document headers. It is common for certain organizations' roles to be absent from the signature box but present in section headings. In some cases, even company names are missing from the signature box and must be retrieved from headers.

    \item \textit{Full Document}: Locations are often not explicitly mentioned in the signature box or headers, requiring a broader document search. Occasionally, some organizations (e.g., Guarantors) are not mentioned in either the signature box or the headers, making a full-document search necessary.
    
\end{itemize}

\paragraph{Relation Annotations}

\begin{itemize}[leftmargin=*,nosep]
    
    \item \colorbox[rgb]{0.96,0.71,0.71}{\texttt{Org Name}} \( \rightarrow \) \colorbox[rgb]{0.75,0.71,0.96}{\texttt{Org Role}}: Each organization name is annotated once per document, while its role is annotated as close to the name as possible. If multiple organizations share the same role (e.g., Borrower) and it is mentioned collectively in the signature box, the role is annotated once and linked to all relevant organizations. Otherwise, each occurrence of a role is annotated separately and linked to the corresponding organization.

    \item \colorbox[rgb]{0.75,0.71,0.96}{\texttt{Org Role}} \( \rightarrow \)  \colorbox[rgb]{0.71,0.91,0.71}{\texttt{Org Sub-Role}}: If an organization has a specific sub-role (e.g., Parent Borrower), its role is annotated separately from others, and the sub-role is linked to it. The connection follows a hierarchical structure, linking the organization to its role, then the role to its sub-role.

    \item \colorbox[rgb]{0.96,0.71,0.71}{\texttt{Org Name}} \( \rightarrow \)  \colorbox[rgb]{0.95,0.95,0.78}{\texttt{Person Name}}: Each person’s name is annotated once per document and linked to a single organization (see \ref{subapp:main_edge_cases} for special scenarios).

    \item \colorbox[rgb]{0.95,0.95,0.78}{\texttt{Person Name}} \( \rightarrow \)  \colorbox[rgb]{0.88,0.76,0.91}{\texttt{Person Position}}: A person’s position is annotated as close to their name as possible. Individuals may hold multiple positions, which are all linked to the same person annotation. In some cases, position mentions may be missing (see Figure \ref{fig:address_person} and \ref{subapp:main_edge_cases}).

    \item \colorbox[rgb]{0.96,0.71,0.71}{\texttt{Org Name}} \( \rightarrow \) \colorbox[rgb]{0.97,0.90,0.71}{\texttt{Location}}: Locations are annotated once per document and linked to the relevant organization.

    \item \colorbox[rgb]{0.97,0.90,0.71}{\texttt{Location}} \( \rightarrow \)  \colorbox[rgb]{0.72,0.90,0.97}{\texttt{Location Type}}: Each location has exactly one type, which is annotated as close to the location name as possible. Location types are often missing, and in such cases, no annotation is applied (see Fig \ref{fig:address_person}).

\end{itemize}

\paragraph{Continuation Links} In cases where the text to be annotated is split across multiple lines or interrupted by symbols, preventing a single coherent annotation, we annotate each segment separately and link them sequentially: the first to the second, the second to the third, and so on. This issue is particularly common for Locations, where addresses or names are fragmented (see Figure \ref{fig:address_person}). Since our annotation structure does not include relations between labels of the same type, these links are easily identifiable. During post-processing, entities connected by continuation links are automatically merged with spaces into a single coherent label.

\subsection{Main Edge Cases}
\label{subapp:main_edge_cases}

\begin{figure*}[htbp]
    \centering
    \begin{subfigure}[b]{0.235\textwidth}
        \centering
        \includegraphics[width=\linewidth]{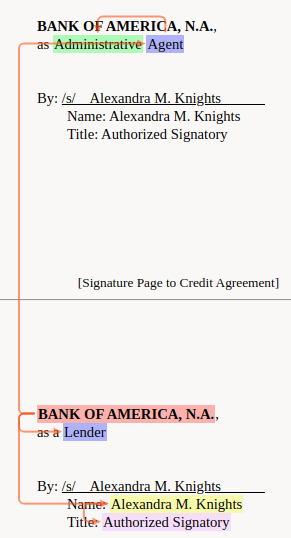}
        \caption{General and most common annotation in the signature box.}
        \label{fig:unique_entity_annotation}
    \end{subfigure}
    \hfill
    \begin{subfigure}[b]{0.375\textwidth}
        \centering
        \includegraphics[width=\linewidth]{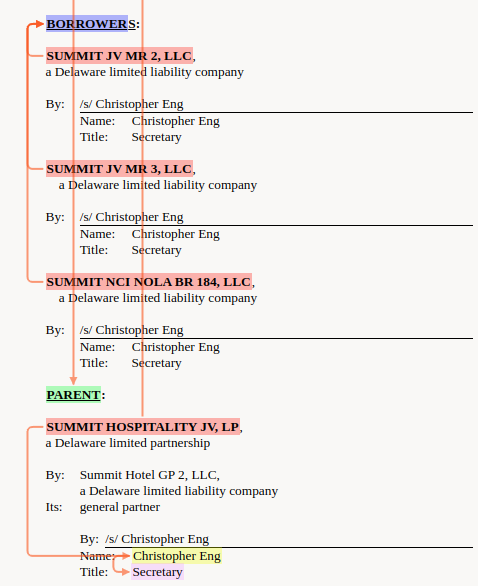}
        \caption{Multiple organizations with similar name being borrowers.}
        \label{fig:same_entity_mult_names_bor}
    \end{subfigure}
    \hfill
    \begin{subfigure}[b]{0.355\textwidth}
        \centering
        \includegraphics[width=\linewidth]{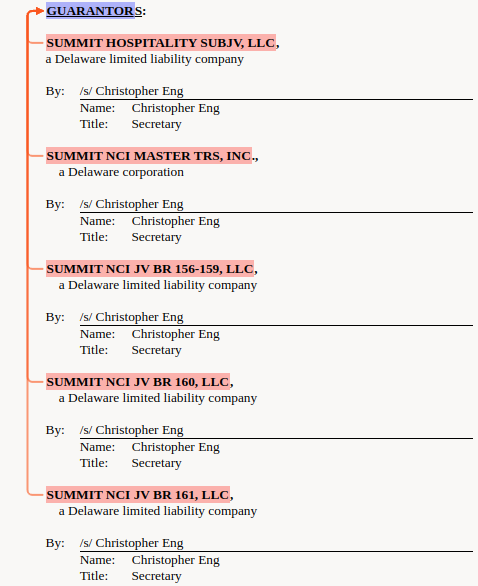}
        \caption{Multiple organizations with similar name being guarantors in addition to being borrowers.}
        \label{fig:same_entity_mult_names_guar}
    \end{subfigure}

    \vspace{1em}
    
    \begin{subfigure}[b]{0.34\textwidth}
        \centering
        \includegraphics[width=\linewidth]{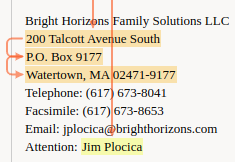}
        \caption{Location without type, person without position and continuation links.}
        \label{fig:address_person}
    \end{subfigure}
    \hfill
    \begin{subfigure}[b]{0.29\textwidth}
        \centering
        \includegraphics[width=\linewidth]{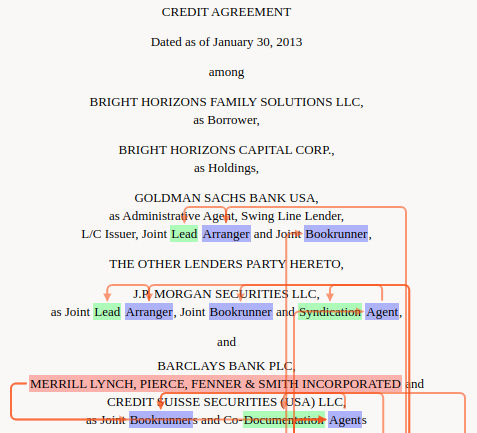}
        \caption{Annotations in the header and joint roles.}
        \label{fig:joint_co}
    \end{subfigure}
    \hfill
    \begin{subfigure}[b]{0.33\textwidth}
        \centering
        \includegraphics[width=\linewidth]{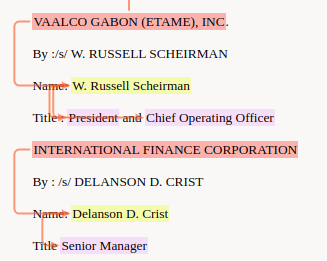}
        \caption{Only 2 organizations, not roles in the signature box or header.}
        \label{fig:Small_Agreement_sign}
    \end{subfigure}

    \vspace{1em}
    
    \begin{subfigure}[b]{\textwidth}
        \centering
        \includegraphics[width=\linewidth]{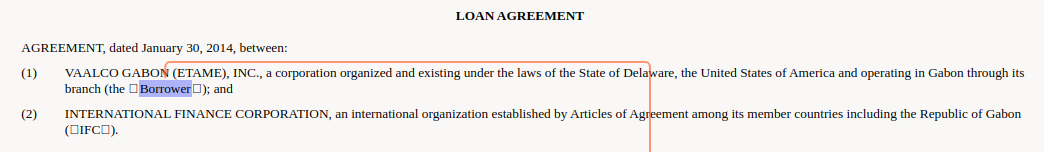}
        \caption{Only 2 organizations, only borrower's role is mentioned.}
        \label{fig:Small_Agreement_role}
    \end{subfigure}

    \caption{Representative Annotation Scenarios from Label Studio.}
    \label{fig:annotation_guide}
\end{figure*}

\paragraph{Parent Borrower and Related Entities} Some agreements list multiple entities as borrowers and guarantors, often sharing highly similar names (e.g., Summit Hospitality JV, LP, Summit JV MR 2, LLC in Figures \ref{fig:same_entity_mult_names_bor} and \ref{fig:same_entity_mult_names_guar}), suggesting a structured division of what may functionally be a single entity. In such cases, one entity is designated as the \texttt{Parent Borrower}, while the others are categorized under \texttt{Borrowers} and \texttt{Guarantors}. The same individual typically signs on behalf of all these entities, but following our general rule, \textit{we annotate a person once} and link them to the \texttt{Parent Borrower}. Additionally, since the Parent Borrower’s role is often labeled simply as \texttt{Parent} in the signature box, while other borrowers explicitly receive the \texttt{Borrower} role, we annotate the closest explicit mention of \texttt{Borrower} elsewhere in the document to ensure accurate labeling.

\paragraph{Missing Person Position}In some cases, individuals are mentioned within address sections rather than as signatories or officials, likely indicating their responsibility for a specific location. These individuals often lack an explicitly stated position. As shown in Figure \ref{fig:address_person}, Jim Plocica appears in the address block without a corresponding title. When this occurs, we annotate the Person Name but leave the Person Position unannotated.

Additionally, this example highlights several annotation conventions:

\begin{itemize}[leftmargin=*,nosep]
    
    \item The Organization Name (Bright Horizons Family Solutions LLC) is omitted here because it is annotated in the signature box (once per document). The links to the Location and Person Name originate from that primary annotation.

    \item The Location spans multiple lines, requiring the use of Continuation Links (described in Section \ref{subapp:general_rules}) to maintain coherence.

    \item No Location Type is explicitly mentioned, which is a common occurrence in such address sections.

\end{itemize}

\paragraph{Shared Roles and Prefixes} When multiple organizations share a role, they are often referred to collectively using prefixes like Joint or Co- (e.g., Joint Bookrunner or Co-Documentation Agent). However, our annotation framework requires capturing role information at the individual organization level. Since a single entity cannot hold a joint role independently, we do not annotate these prefixes. Instead, we annotate only the core role (Bookrunner, Documentation Agent, etc.) for each company separately, as illustrated in Figure \ref{fig:joint_co}).

Additionally, in this example, the annotations are sourced from the header rather than the signature box because Merrill Lynch, Pierce, Fenner \& Smith Incorporated is absent from the signature box, and for other companies, only their primary role (Lender) is mentioned there.

\paragraph{Smallest Document} One of the extracted credit agreements involves only 2 organizations, and their roles are mentioned neither in the signature box nor in the header (Figure \ref{fig:Small_Agreement_sign}). Moreover, the role of the lender is not mentioned in the document, while the role of the borrower is mentioned once (Figure \ref{fig:Small_Agreement_role}).

\subsection{Annotation Process Details}
\label{subapp:annotation_process_details}

We annotated the extracted credit agreements using Label-Studio\footnote{\url{https://github.com/HumanSignal/label-studio}}, supporting entity linking and relation extraction for \texttt{.html} files. The annotation was carried out by annotators with backgrounds in Computer Science or Mathematics, under the direct supervision of a Master of Finance and a PhD student advised by a Chair Professor of Finance, also a co-author. The annotation followed a structured, iterative approach to ensure consistency and resolve edge cases. The team first established label definitions and completed an initial annotation round. The results were reviewed to identify and discuss edge cases (Appendix \ref{subapp:main_edge_cases}), which led to refinements in the annotation guidelines and the second round of annotations (inter-annotator agreement was not calculated at this point, as the guidelines were evolving). Following the second annotation round, we constructed knowledge-graph (KG) document representations, sampled graphs, and compared them against the ground-truth annotations. Insights from this qualitative assessment (Appendix \ref{subapp:graph_analysis}) informed a third annotation round, after which we applied an additional cleaning script to the KG representations. 

\subsection{Complexity vs coverage}
\label{subapp:complexity_vs_coverage}

\frameworkshortname{} defines seven entity and six relation types, and this choice reflects a deliberate trade-off between semantic coverage and compositional complexity. Expanding the ontology would increase topical breadth but would not necessarily yield deeper reasoning requirements. Our goal is to evaluate long-context comprehension under settings where information is distributed across distant parts of a document and must be \textbf{\textit{retrieved, composed, and reasoned over}}.

Thus, the benchmark emphasizes complex inference chains (multi-hop reasoning, set operations, and answer plurality) over the schema coverage. Answering questions from our dataset requires linking dispersed mentions, disambiguating roles, and integrating partial evidence: operations that jointly test both retrieval and reasoning. This structure allows us to probe failure modes of long-context models in a controlled environment, balancing interpretability, scalability, and difficulty without relying on full document coverage.

\clearpage

\section{Knowledge Graph Construction Details}
\label{app:knowledge_graph_construction}

\paragraph{Ontology Layer} The ontology layer defines a class hierarchy that structures entities in the knowledge graph. We organize annotated entity types into three top-level classes — \texttt{Person}, \texttt{Organization}, and \texttt{Location} — based on their functional roles in financial agreements. These are connected to their respective sub-categories through explicit \texttt{Sub-class} relationships, capturing hierarchical distinctions. For example, \texttt{Organization} includes roles such as \texttt{Lender}, \texttt{Agent}, and \texttt{L/C Issuer}, while \texttt{Person} includes positions like \texttt{Vice President} and \texttt{Treasurer}. To further refine responsibilities, we introduce a second level of subclasses via \texttt{Sub-Roles}, such as dividing \texttt{Agent} into \texttt{Administrative Agent} and \texttt{Documentation Agent}. Only these role and position types appear as classes in the ontology layer; the actual named entities (e.g., ``Jane Smith", ``ABCDE Inc.") are treated as instances in the data layer. This abstraction supports generalization across documents and enables scalable, template-based QA pairs extraction.

\paragraph{Data Layer} Complementing the ontology layer, the data layer grounds the class structure in real-world instances and their relationships. Entities such as \texttt{ABCDE Inc.}, \texttt{John Doe}, and \texttt{10 Peachtree} instantiate ontology classes like \texttt{Organization}, \texttt{Person}, and \texttt{Location} via \texttt{Instance Of} links. These instances are further connected through \texttt{Inter-Instance} relations (e.g., \texttt{hasEmployee}, \texttt{isLocationOf}), capturing document-specific associations such as employment and location.

\paragraph{Ontology and Data Scope} While each credit agreement yields its own data layer containing document-specific instances and relationships, the ontology layer is shared across all documents. This unified ontology defines a consistent schema for entity types and roles, enabling systematic alignment across diverse agreements.

\paragraph{Question Construction with SPARQL} SPARQL queries enable automated, structured QA extraction from our knowledge graph. This approach dynamically adapts to diverse financial agreements by leveraging ontology. The query in Figure \ref{fig:sparql_query} retrieves individuals and their positions using the \texttt{isInstanceOf} relationship, while \texttt{GROUP\_CONCAT} consolidates multiple positions per individual for comprehensive QA pairs extraction. 

\begin{figure}[h]
    \centering
    
    \begin{lstlisting}
SELECT DISTINCT ?person (GROUP_CONCAT(?position; separator="|") as ?positions)
WHERE {
    ?person a <http://example.org/base/Person> ;
           <http://example.org/isInstanceOf/> ?position .
    FILTER(STRSTARTS(STR(?position), "http://example.org/person_position/"))
}
GROUP BY ?person
    \end{lstlisting}
    \caption{SPARQL Query Example for Structured QA Extraction.}
    \label{fig:sparql_query}
\end{figure}

\subsection{Qualitative Analysis of the Constructed KGs}
\label{subapp:graph_analysis}
After generating question-answer pairs from the extracted graphs, we observed two common issues affecting the quality and consistency of the QAs.

\begin{itemize}[leftmargin=*,nosep]
    \item \textbf{Duplicate entries:} Multiple questions or answers representing the same entity or relation, differing only by capitalization, whitespace, punctuation, or pluralization.  
    \item \textbf{Noisy text:} URL-encoded characters (e.g., \texttt{\%26}, \texttt{\%20}) and miscellaneous special characters were present in questions and answers.  
\end{itemize}

To address these issues, we applied a structured cleaning procedure \textbf{\textit{after QA generation}}, ensuring the graphs themselves were not altered but the extracted QAs were reliable:

\begin{itemize}[leftmargin=*,nosep]
    \item \textbf{Text normalization}: 
    \begin{itemize}[nosep]
        \item Convert all text to lowercase.
        \item Remove URL-encoded sequences and special characters, preserving commas for multiple answers.
        \item Collapse multiple whitespace characters.
    \end{itemize}

    \item \textbf{Duplicate answer removal}:
    \begin{itemize}[nosep]
        \item Split multi-answer entries by commas and clean each entry.
        \item Normalize text (lowercase, remove spaces) and reduce plural forms to singular where appropriate.
        \item Remove duplicates while preserving the first occurrence.
        \item Log all removed duplicates, including document identifiers, for traceability.
    \end{itemize}
\end{itemize}

\section{All templates}
\label{app:all_templates}

This appendix contains the defined question templates.

\begin{table*}[htbp]
    \caption{Level 1 Question Templates with Counts.}
    \centering
    \resizebox{\linewidth}{!}{
    \begin{tabular}{p{0.38\textwidth}p{0.25\textwidth}p{0.03\textwidth}p{0.03\textwidth}p{0.03\textwidth}p{0.10\textwidth}}
    \toprule
    \textbf{Template} & \textbf{Based On} & \textbf{P} & \textbf{H} & \textbf{\#SO} & \textbf{Count} \\
    \midrule
    What is the position of [Person Name]? [if one] &  & 0 & 1 & 0 & 1252 \\
    In what organization does [Person Name] work? &  & 0 & 1 & 0 & 1288 \\
    Who is the representative of [Org Name]? [if one] &  & 0 & 1 & 0 & 1358 \\
    What is the role of [Org Name] in the agreement? [if one] &  & 0 & 1 & 0 & 845 \\
    What company is the [Org Role (+ Sub-Role)] in the agreement? [if one] &  & 0 & 1 & 0 & 444 \\
    What is the location of [Org Name]? [if one] &  & 0 & 1 & 0 & 790 \\
    Which company is associated with [Location]? &  & 0 & 1 & 0 & 530 \\
    What type of location is [Location] (e.g., Headquarters, Trade Operations, etc.)? [if one] &  & 0 & 1 & 0 & 43 \\
    \bottomrule
    \end{tabular}
    }
    \label{tab:all_templatex_L1}
\end{table*}

\begin{table*}[htbp]
    \caption{Level 2 Question Templates with Counts.}
    \centering
    \resizebox{\linewidth}{!}{
    \begin{tabular}{p{0.38\textwidth}p{0.25\textwidth}p{0.03\textwidth}p{0.03\textwidth}p{0.03\textwidth}p{0.10\textwidth}}
    \toprule
    \textbf{Template} & \textbf{Based On} & \textbf{P} & \textbf{H} & \textbf{\#SO} & \textbf{Count} \\
    \midrule
    Who is the [Person Position] of [Org Name]? [if one] &  & 0 & 2 & 0 & 2274 \\
    What is the position held by both [Person Name 1] and [Person Name 2]? [if one] & What is the position of [Person Name]? [if one] & 0 & 1 & 1 & 1612 \\
    What is the role in the agreement of the company where [Person Name] is employed? [if one] & What is the role of [Org Name] in the agreement? [if one] & 0 & 2 & 0 & 613 \\
    What are the roles of [Org Name] in the agreement? & What is the role of [Org Name] in the agreement? [if one] & 1 & 1 & 0 & 409 \\
    What companies are the [Org Role (+ Sub-Role)] in the agreement? & What company is the [Org Role (+ Sub-Role)] in the agreement? [if one] & 1 & 1 & 0 & 304 \\
    What role do both [Org Name 1] and [Org Name 2] have in the agreement? [if one] & What is the role of [Org Name] in the agreement? [if one] & 0 & 1 & 1 & 257 \\
    What is the role in the agreement of the company associated with [Location]? [if one] & What is the role of [Org Name] in the agreement? [if one] & 0 & 2 & 0 & 180 \\
    Who are the representatives of [Org Name]? & Who is the representative of [Org Name]? [if one] & 1 & 1 & 0 & 160 \\
    What are the positions of [Person Name]? & What is the position of [Person Name]? [if one] & 1 & 1 & 0 & 136 \\
    What are the locations of [Org Name]? & What is the location of [Org Name]? [if one] & 1 & 1 & 0 & 91 \\
    In what organizations does [Person Name] work? &  In what organization does [Person Name] work?  & 1 & 1 & 0 & 90 \\
    What is the [Location Type] office of [Org Name]? &  & 0 & 2 & 0 & 70 \\
    What types of location is [Location] (e.g., Headquarters, Trade Operations, etc.)? & What type of location is [Location] (e.g., Headquarters, Trade Operations, etc.)? [if one] & 1 & 1 & 0 & 11 \\
    \bottomrule
    \end{tabular}
    }
    \label{tab:all_templatex_L2}
\end{table*}

\clearpage

\begin{table*}[htbp]
    \caption{Level 3 Question Templates with Counts.}
    \centering
    \resizebox{\linewidth}{!}{
    \begin{tabular}{p{0.42\textwidth}p{0.40\textwidth}p{0.03\textwidth}p{0.03\textwidth}p{0.03\textwidth}p{0.05\textwidth}}
    \toprule
    \textbf{Template} & \textbf{Based On} & \textbf{P} & \textbf{H} & \textbf{\#SO} & \textbf{Count} \\
    \midrule
    What are the positions held by both [Person Name 1] and [Person Name 2]? & What are the positions of [Person Name]? & 1 & 1 & 1 & 3 \\
    What is the position held by [Person Name 1] but not by [Person Name 2]? [if one] & What are the positions held by both [Person Name 1] and [Person Name 2]? & 0 & 1 & 2 & 8 \\
    \midrule
    What roles do both [Org Name 1] and [Org Name 2] have in the agreement? & What are the roles of [Org Name] in the agreement? & 1 & 1 & 1 & 92 \\
    What role does [Org Name 1] have in the agreement which is not the role of [Org Name 2]? [if one] & What roles do both [Org Name 1] and [Org Name 2] have in the agreement? & 0 & 1 & 2 & 278 \\
    \midrule
    What company is the [Org Role (+ Sub-Role) 1] but not the [Org Role (+ Sub-Role) 2] in the agreement? [if one] & What companies are both the [Org Role (+ Sub-Role) 1] and [Org Role (+ Sub-Role) 2] in the agreement? & 0 & 1 & 2 & 56 \\
    \midrule
    Who are the [Person Position]s of [Org Name]? & Who is the [Person Position] of [Org Name]? [if one] & 1 & 2 & 0 & 57 \\
    Who is the [Person Position 1] and [Person Position 2] of [Org Name]? [if one] & Who is the [Person Position] of [Org Name]? [if one] & 0 & 2 & 1 & 1346 \\
    \midrule
    What are the roles in the agreement of the company where [Person Name] is employed? & What is the role in the agreement of the company where [Person Name] is employed? [if one] & 1 & 2 & 0 & 318 \\
    What are the roles in the agreement of the company associated with [Location]? & What is the role in the agreement of the company associated with [Location]? [if one] & 1 & 2 & 0 & 169 \\
    \midrule
    Who is the [Person Position] of the company which is the [Org Role(-s) (+ Sub-Role(-s))] in the agreement? [if one, and the company should be uniquely identifiable] & Who is the [Person Position] of [Org Name]? [if one] & 0 & 3 & 0 & 262 \\
    Who is the [Person Position] of the company associated with [Location]? [if one] & Who is the [Person Position] of [Org Name]? [if one] & 0 & 3 & 0 & 616 \\
    Who is the [Person Position] of the company where [Person Name] is employed? [if one] & Who is the [Person Position] of [Org Name]? [if one] & 0 & 3 & 0 & 289 \\
    \midrule
    What is the address of [Location Type] of the company which is the [Org Role(-s) (+ Sub-Role(-s))] in the agreement? [if one, and the company should be uniquely identifiable] & What is the [Location Type] office of [Org Name]? & 0 & 3 & 0 & 13 \\
    What is the address of [Location Type] of the company where [Person Name] is employed? & What is the [Location Type] office of [Org Name]? & 0 & 3 & 0 & 75 \\
    Who is the [Person Position] of the company where [Person Name] is employed? [if one] & Who is the [Person Position] of [Org Name]? [if one] & 0 & 3 & 0 & 289 \\
    \bottomrule
    \end{tabular}
    }
    \label{tab:all_templatex_L3}
\end{table*}

\clearpage

\begin{table*}[htbp]
    \caption{Level 4 Question Templates, part 1.}
    \centering
    \resizebox{\linewidth}{!}{
    \begin{tabular}{p{0.40\textwidth}p{0.40\textwidth}p{0.03\textwidth}p{0.03\textwidth}p{0.03\textwidth}p{0.05\textwidth}}
    \toprule
    \textbf{Template} & \textbf{Based On} & \textbf{P} & \textbf{H} & \textbf{\#SO} & \textbf{Count} \\
    \midrule
    What are the positions held by [Person Name 1] but not by [Person Name 2]? & What are the positions held by both [Person Name 1] and [Person Name 2]?
    What is the position held by [Person Name 1] but not by [Person Name 2]? [if one] & 1 & 1 & 2 & 2 \\
    What roles does [Org Name 1] have in the agreement which are not the roles of [Org Name 2]? & What roles do both [Org Name 1] and [Org Name 2] have in the agreement?
    What role does [Org Name 1] have in the agreement which is not the role of [Org Name 2]? [if one] & 1 & 1 & 2 & 134 \\
    What companies are the [Org Role (+ Sub-Role) 1] but not the [Org Role (+ Sub-Role) 2] in the agreement? & What companies are both the [Org Role (+ Sub-Role) 1] and [Org Role (+ Sub-Role) 2] in the agreement?
    What company is the [Org Role (+ Sub-Role) 1] but not the [Org Role (+ Sub-Role) 2] in the agreement? [if one] & 1 & 1 & 2 & 124 \\
    \midrule
    What is the position held by [Person Name 1] but not by [Person Name 2] or [Person Name 3]? [if one] & What is the position held by [Person Name 1] but not by [Person Name 2]? [if one] & 0 & 1 & 3 & 50 \\
    What is the position held by [Person Name 1] and [Person Name 2] but not by [Person Name 3]? [if one] & What is the position held by [Person Name 1] but not by [Person Name 2]? [if one] & 0 & 1 & 3 & 4 \\
    What role do [Org Name 1] and [Org Name 2] have in the agreement which is not the role of [Org Name 3]? [if one] & What role does [Org Name 1] have in the agreement which is not the role of [Org Name 2]? [if one] & 0 & 1 & 3 & 0 \\
    What role does [Org Name 1] have in the agreement which is not the role of [Org Name 2] or [Org Name 3]? [if one] & What role does [Org Name 1] have in the agreement which is not the role of [Org Name 2]? [if one] & 0 & 1 & 3 & 0 \\
    What company is the [Org Role (+ Sub-Role) 1] and [Org Role (+ Sub-Role) 2] but not the [Org Role (+ Sub-Role) 3] in the agreement? [if one] & What company is the [Org Role (+ Sub-Role) 1] but not the [Org Role (+ Sub-Role) 2] in the agreement? [if one] & 0 & 1 & 3 & 1553 \\
    What company is the [Org Role (+ Sub-Role) 1] but not the [Org Role (+ Sub-Role) 2] [Org Role (+ Sub-Role) 3] in the agreement? [if one] & What company is the [Org Role (+ Sub-Role) 1] but not the [Org Role (+ Sub-Role) 2] in the agreement? [if one] & 0 & 1 & 3 & 816 \\
    \bottomrule
    \end{tabular}
    }
    \label{tab:all_templatex_L4_1}
\end{table*}

\clearpage

\begin{table*}[htbp]
    \caption{Level 4 Question Templates, part 2.}
    \centering
    \resizebox{\textwidth}{!}{
    \begin{tabular}{p{0.40\textwidth}p{0.35\textwidth}p{0.03\textwidth}p{0.03\textwidth}p{0.03\textwidth}p{0.05\textwidth}}
    \toprule
    \textbf{Template} & \textbf{Based On} & \textbf{P} & \textbf{H} & \textbf{\#SO} & \textbf{Count} \\
    \midrule
    Who are the both [Person Position 1]s and [Person Position 2]s of [Org Name]? & Who are the [Person Position]s of [Org Name]?
    Who is the [Person Position 1] and [Person Position 2] of [Org Name]? [if one] & 1 & 2 & 1 & 0 \\
    \midrule
    Who are the [Person Position]s of the company which is the [Org Role(-s) (+ Sub-Role(-s))] in the agreement? [the company should be uniquely identifiable] & Who is the [Person Position] of the company which is the [Org Role(-s) (+ Sub-Role(-s))] in the agreement? [if one, and the company should be uniquely identifiable] & 1 & 3 & 0 & 40 \\
    Who are the [Person Position]s of the company associated with [Location]? & Who is the [Person Position] of the company associated with [Location]? [if one] & 1 & 3 & 0 & 22 \\
    Who are the [Person Position]s of the company associated where [Person Name] is employed? & Who is the [Person Position] of the company associated where [Person Name] is employed? [if one] & 1 & 3 & 0 & 5 \\
    \midrule
    Who is the [Person Position] of the company which is both the [Org Role(-s) (+ Sub-Role(-s)) 1] and the [Org Role(-s) (+ Sub-Role(-s)) 2] in the agreement? [if one, and the company should be uniquely identifiable] & Who is the [Person Position] of the company which is the [Org Role(-s) (+ Sub-Role(-s))] in the agreement? [if one, and the company should be uniquely identifiable] & 0 & 3 & 1 & 187 \\
    Who is both the [Person Position 1] and [Person Position 2] of the company which is the [Org Role(-s) (+ Sub-Role(-s))] in the agreement? [if one, and the company should be uniquely identifiable] & Who is the [Person Position] of the company which is the [Org Role(-s) (+ Sub-Role(-s))] in the agreement? [if one, and the company should be uniquely identifiable] & 0 & 3 & 1 & 61 \\
    Who is both the [Person Position 1] and [Person Position 2] of the company associated with [Location]? [if one] & Who is the [Person Position] of the company associated with [Location]? [if one] & 0 & 3 & 1 & 59 \\
    Who is both the [Person Position 1] and [Person Position 2] of the company associated where [Person Name] is employed? [if one] & Who is the [Person Position] of the company associated where [Person Name] is employed? [if one] & 0 & 3 & 1 & 23 \\
    What is the address of the [Location Type] office of the company which is both the [Org Role(-s) (+ Sub-Role(-s)) 1] and the [Org Role(-s) (+ Sub-Role(-s)) 2] in the agreement? [if one, and the company should be uniquely identifiable] & What is the [Location Type] office of the company which is the [Org Role(-s) (+ Sub-Role(-s))] in the agreement? [if one, and the company should be uniquely identifiable] & 0 & 3 & 1 & 14 \\
    \bottomrule
    \end{tabular}
    }
    \label{tab:all_templatex_L4_2}
\end{table*}

\clearpage

\begin{table*}[htbp]
    \caption{Level 5 Question Templates, part 1.}
    \centering
    \resizebox{\linewidth}{!}{
    \begin{tabular}{p{0.40\textwidth}p{0.40\textwidth}p{0.03\textwidth}p{0.03\textwidth}p{0.03\textwidth}p{0.05\textwidth}}
    \toprule
    \textbf{Template} & \textbf{Based On} & \textbf{P} & \textbf{H} & \textbf{\#SO} & \textbf{Count} \\
    \midrule
    What are the positions held by [Person Name 1] but not by [Person Name 2] or [Person Name 3]? & What are the positions held by [Person Name 1] but not by [Person Name 2]?
    What is the position held by [Person Name 1] but not by [Person Name 2] or [Person Name 3]? [if one] & 1 & 1 & 3 & 8 \\
    What are the positions held by [Person Name 1] and [Person Name 2] but not by [Person Name 3]? & What are the positions held by [Person Name 1] but not by [Person Name 2]?
    What is the position held by [Person Name 1] and [Person Name 2] but not by [Person Name 3]? [if one] & 1 & 1 & 3 & 49 \\
    What roles do [Org Name 1] and [Org Name 2] have in the agreement which are not the roles of [Org Name 3]? & What roles does [Org Name 1] have in the agreement which are not the roles of [Org Name 2]?
    What role do [Org Name 1] and [Org Name 2] have in the agreement which is not the role of [Org Name 3]? [if one] & 1 & 1 & 3 & 0 \\
    What roles does [Org Name 1] have in the agreement which are not the roles of [Org Name 2] or [Org Name 3]? & What roles does [Org Name 1] have in the agreement which are not the roles of [Org Name 2]?
    What role does [Org Name 1] have in the agreement which is not the role of [Org Name 2] or [Org Name 3]? [if one] & 1 & 1 & 3 & 0 \\
    What companies are the [Org Role (+ Sub-Role) 1] and [Org Role (+ Sub-Role) 2] but not the [Org Role (+ Sub-Role) 3] in the agreement? & What companies are the [Org Role (+ Sub-Role) 1] but not the [Org Role (+ Sub-Role) 2] in the agreement?
    What company is the [Org Role (+ Sub-Role) 1] and [Org Role (+ Sub-Role) 2] but not the [Org Role (+ Sub-Role) 3] in the agreement? [if one] & 1 & 1 & 3 & 122 \\
    What companies are the [Org Role (+ Sub-Role) 1] but not the [Org Role (+ Sub-Role) 2] [Org Role (+ Sub-Role) 3] in the agreement? [if one] & What companies are the [Org Role (+ Sub-Role) 1] but not the [Org Role (+ Sub-Role) 2] in the agreement?
    What company is the [Org Role (+ Sub-Role) 1] but not the [Org Role (+ Sub-Role) 2] [Org Role (+ Sub-Role) 3] in the agreement? [if one] & 1 & 1 & 3 & 0 \\
    \bottomrule
    \end{tabular}
    }
    \label{tab:all_templatex_L5_1}
\end{table*}

\clearpage

\begin{table*}[htbp]
    \caption{Level 5 Question Templates, part 2.}
    \centering
    \resizebox{\linewidth}{!}{
    \begin{tabular}{p{0.40\textwidth}p{0.40\textwidth}p{0.03\textwidth}p{0.03\textwidth}p{0.03\textwidth}p{0.05\textwidth}}
    \toprule
    \textbf{Template} & \textbf{Based On} & \textbf{P} & \textbf{H} & \textbf{\#SO} & \textbf{Count} \\
    \midrule
    Who are the [Person Position]s of the company which is both the [Org Role(-s) (+ Sub-Role(-s)) 1] and the [Org Role(-s) (+ Sub-Role(-s)) 2] in the agreement? [the company should be uniquely identifiable] & Who are the [Person Position]s of the company which is the [Org Role(-s) (+ Sub-Role(-s))] in the agreement? [the company should be uniquely identifiable] Who is the [Person Position] of the company which is both the [Org Role(-s) (+ Sub-Role(-s)) 1] and the [Org Role(-s) (+ Sub-Role(-s)) 2] in the agreement? [if one, and the company should be uniquely identifiable] & 1 & 3 & 1 & 0 \\
    Who are both the [Person Position 1] and [Person Position 2] of the company which is the [Org Role(-s) (+ Sub-Role(-s))] in the agreement? [the company should be uniquely identifiable] & Who are the [Person Position]s of the company which is the [Org Role(-s) (+ Sub-Role(-s))] in the agreement? [the company should be uniquely identifiable] Who is both the [Person Position 1] and [Person Position 2] of the company which is the [Org Role(-s) (+ Sub-Role(-s))] in the agreement? [if one, and the company should be uniquely identifiable] & 1 & 3 & 1 & 0 \\
    Who are both the [Person Position 1] and [Person Position 2] of the company associated with [Location]? & Who are the [Person Position]s of the company associated with [Location]?
    Who is both the [Person Position 1] and [Person Position 2] of the company associated with [Location]? [if one] & 1 & 3 & 1 & 0 \\
    Who are both the [Person Position 1] and [Person Position 2] of the company associated where [Person Name] is employed? & Who are the [Person Position]s of the company associated where [Person Name] is employed? Who is both the [Person Position 1] and [Person Position 2] of the company associated where [Person Name] is employed? [if one] & 1 & 3 & 1 & 0 \\
    \midrule
    Who is the [Person Position 1] but not [Person Position 2] of the company which is the [Org Role(-s) (+ Sub-Role(-s))] in the agreement? [if one, and the company should be uniquely identifiable] & Who is both the [Person Position 1] and [Person Position 2] of the company which is the [Org Role(-s) (+ Sub-Role(-s))] in the agreement? [if one, and the company should be uniquely identifiable] & 0 & 3 & 2 & 2 \\
    Who is the [Person Position 1] but not [Person Position 2] of the company associated with [Location]? [if one] & Who is both the [Person Position 1] and [Person Position 2] of the company associated with [Location]? [if one] & 0 & 3 & 2 & 1 \\
    Who is the [Person Position 1] but not [Person Position 2] of the company associated where [Person Name] is employed? [if one] & Who is both the [Person Position 1] and [Person Position 2] of the company associated where [Person Name] is employed? [if one] & 0 & 3 & 2 & 3 \\
    Who is the [Person Position] of the company which is the [Org Role(-s) (+ Sub-Role(-s)) 1] but not the [Org Role(-s) (+ Sub-Role(-s)) 2] in the agreement? [if one, and the company should be uniquely identifiable] & Who is the [Person Position] of the company which is both the [Org Role(-s) (+ Sub-Role(-s)) 1] and the [Org Role(-s) (+ Sub-Role(-s)) 2] in the agreement? [if one, and the company should be uniquely identifiable] & 0 & 3 & 2 & 508 \\
    What is the [Location Type] office of the company which is both the [Org Role(-s) (+ Sub-Role(-s)) 1] but not the [Org Role(-s) (+ Sub-Role(-s)) 2] in the agreement? [if one, and the company should be uniquely identifiable] & What is the [Location Type] office of the company which is both the [Org Role(-s) (+ Sub-Role(-s)) 1] and the [Org Role(-s) (+ Sub-Role(-s)) 2] in the agreement? [if one, and the company should be uniquely identifiable] & 0 & 3 & 2 & 13 \\
    \bottomrule
    \end{tabular}
    }
    \label{tab:all_templatex_L5_2}
\end{table*}

\clearpage

\section{Evaluation Details}
\label{app:evaluation_details}

\subsection{Prompting Strategy}
\label{subapp:prompting_strategy}

\begin{figure*}[p]
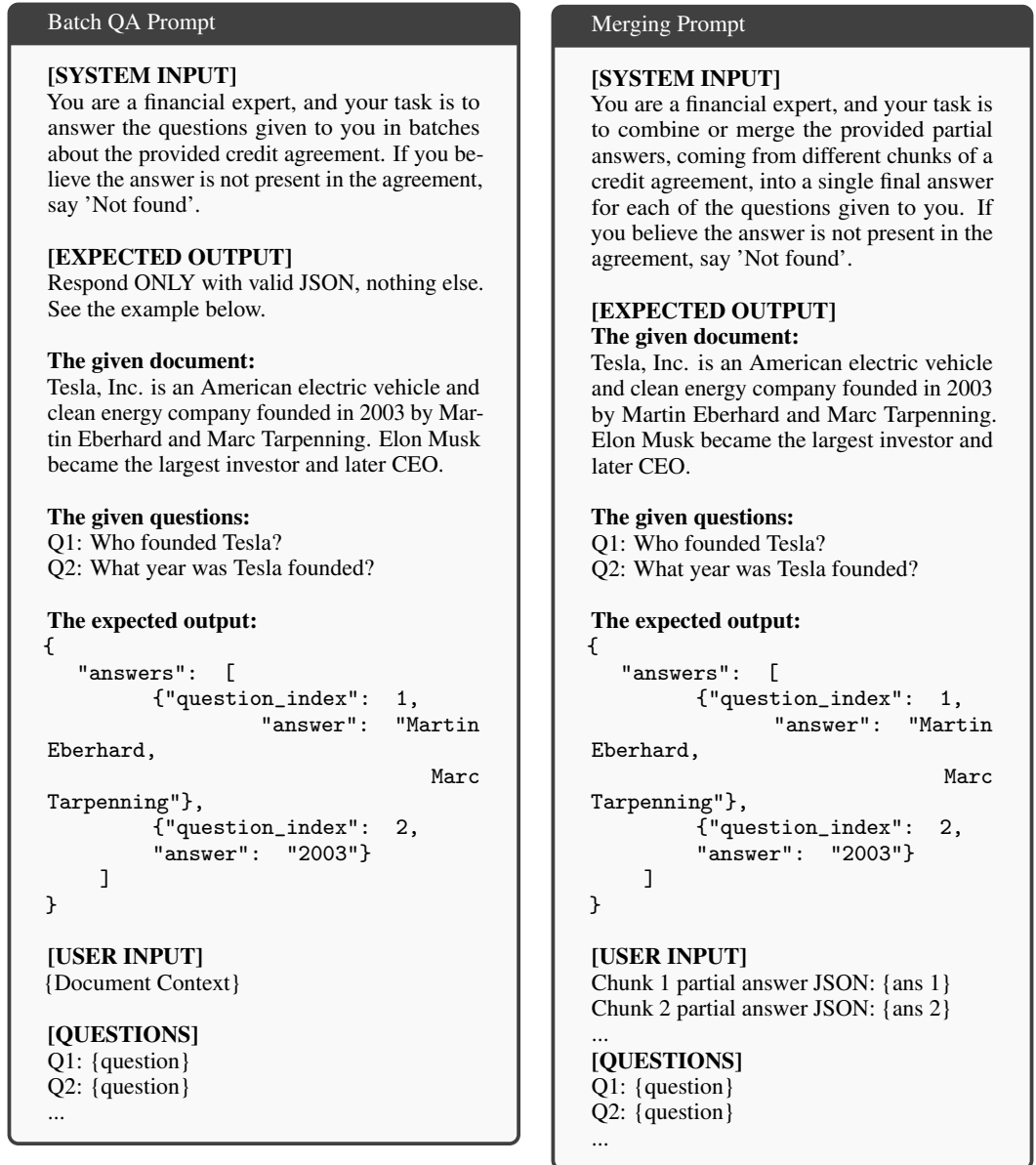

    \begin{subfigure}[b]{0.5\textwidth}
        \centering
        \footnotesize
        \begin{tcolorbox}[colback=gray!5,colframe=black!75,title=Batch QA Prompt]
            \small
            \textbf{[SYSTEM INPUT]} \\
            You are a financial expert, and your task is to answer the questions given to you in batches about the provided credit agreement. If you believe the answer is not present in the agreement, say 'Not found'. \\
            \vspace{0.01cm}
    
            \textbf{[EXPECTED OUTPUT]} \\
            Respond ONLY with valid JSON, nothing else. See the example below.\\
            \vspace{0.01cm}
    
            \textbf{The given document:} \\
            Tesla, Inc. is an American electric vehicle and clean energy company founded in 2003 by Martin Eberhard and Marc Tarpenning. 
            Elon Musk became the largest investor and later CEO. \\
            \vspace{0.01cm}
    
            \textbf{The given questions:} \\
            Q1: Who founded Tesla? \\
            Q2: What year was Tesla founded? \\
            \vspace{0.01cm}
    
            \textbf{The expected output:} \\
            \texttt{\{}
            
            \( \quad \) \texttt{"answers": [} 
            
            \( \qquad \) \( \qquad \) \texttt{\{"question\_index": 1,}
            
            \( \qquad \) \( \qquad \) \texttt{"answer": "Martin Eberhard,}
            
            \( \qquad \) \( \qquad \) \( \qquad \) \( \qquad \) \( \quad \) \texttt{Marc Tarpenning"\},} 
            
            \( \qquad \) \( \qquad \) \texttt{\{"question\_index": 2,}
            
            \( \qquad \) \( \qquad \) \texttt{"answer": "2003"\}}
    
            \( \qquad \) \texttt{]}
            
            \texttt{\}} \\
            \vspace{0.01cm}
    
            \textbf{[USER INPUT]} \\
            \{Document Context\} \\
            \vspace{0.01cm}
    
            \textbf{[QUESTIONS]} \\
            Q1: \{question\} \\
            Q2: \{question\} \\
            ...
        \end{tcolorbox}
        \caption{Example of a batch prompt combining up to 50 questions with a single document chunk. The LLM is instructed to return only JSON-formatted answers, each linked to a question index.}
        \label{fig:batch_qa_prompt}
    \end{subfigure}
    \hfill
    \begin{subfigure}[b]{0.47\textwidth}
        \centering
        \footnotesize
        \begin{tcolorbox}[colback=gray!5,colframe=black!75,title=Merging Prompt]
            \small
            \textbf{[SYSTEM INPUT]} \\
            You are a financial expert, and your task is to combine or merge the provided partial answers, coming from different chunks of a credit agreement, into a single final answer for each of the questions given to you. If you believe the answer is not present in the agreement, say 'Not found'. \\
            \vspace{0.01cm}
    
            \textbf{[EXPECTED OUTPUT]} \\
            \textbf{The given document:} \\
            Tesla, Inc. is an American electric vehicle and clean energy company founded in 2003 by Martin Eberhard and Marc Tarpenning. Elon Musk became the largest investor and later CEO. \\
            
            \textbf{The given questions:} \\
            Q1: Who founded Tesla? \\
            Q2: What year was Tesla founded? \\
            
            \textbf{The expected output:} \\
            \texttt{\{}
            
            \( \quad \) \texttt{"answers": [} 
            
            \( \qquad \) \( \qquad \) \texttt{\{"question\_index": 1,}
            
            \( \qquad \) \( \qquad \) \texttt{"answer": "Martin Eberhard,}
            
            \( \qquad \) \( \qquad \) \( \qquad \) \( \qquad \) \( \quad \) \texttt{Marc Tarpenning"\},} 
            
            \( \qquad \) \( \qquad \) \texttt{\{"question\_index": 2,}
            
            \( \qquad \) \( \qquad \) \texttt{"answer": "2003"\}}
    
            \( \qquad \) \texttt{]}
            
            \texttt{\}} \\
            \vspace{0.01cm}
    
            \textbf{[USER INPUT]} \\
            Chunk 1 partial answer JSON: \{ans 1\} \\
            Chunk 2 partial answer JSON: \{ans 2\} \\
            ...
            \vspace{0.01cm}
    
            \textbf{[QUESTIONS]} \\
            Q1: \{question\} \\
            Q2: \{question\} \\
            ...
        \end{tcolorbox}
        \caption{ Example of a merging prompt that consolidates partial answers from multiple document chunks into one final JSON-formatted answer per question.}
        \label{fig:merging_prompt}
    \end{subfigure}

    \caption{Two-stage prompting pipeline. Long documents are chunked if needed and paired with question batches (\( \leqslant \! 50 \)). A structured one-shot prompt is used per chunk. For multi-chunk documents, a merging prompt consolidates partial answers.}
    \label{fig:prompts}
\end{figure*}

We implement a two-stage prompting pipeline for QA inference over long documents. In the first stage, the document is either processed \textbf{in full or split into chunks} (depending on the context limit of the LLM), and \textbf{each chunk is paired with a batch of up to 50 questions}. We then apply a one-shot, instruction-based prompt (Figure \ref{fig:batch_qa_prompt}) that instructs the LLM to act as a financial expert and return an array of answers -- either grounded in the document or explicitly stating ``Not found" -- in valid JSON format. The effect of chunk-based evaluation is discussed in Appendix \ref{subapp:chunk_based_evaluation}.

The expected format is enforced both through the prompt instructions and through structured parsing logic. The parser first attempts to \textbf{extract a JSON object from the model’s response}, supporting both raw JSON and code-fenced output. If parsing fails, the system falls back to extracting answers from list-formatted outputs or reattempts the query, up to a maximum number of retries. All models are queried with temperature=0.0.

In the second stage, for questions where answers may be distributed across multiple chunks, a merging prompt is issued (Figure \ref{fig:merging_prompt}). This prompt \textbf{consolidates partial answers from earlier responses} and asks the model to output a single final answer per question, again using the same strict JSON format. This strategy allows the model to incorporate information across multiple views of the document.

Regarding resources, as mentioned in Section \ref{subsec:baselines}, for open-weight models we are using Together.AI for some of the models. For the other open-weight models, we used 1 H200 GPU.

\subsection{Evaluation Metrics}
\label{subapp:metrics}

We employ four complementary metrics to evaluate model responses (Table \ref{tab:full_results}).

\begin{itemize}[leftmargin=*,nosep]

    \item \textbf{Word-Level F1 Score} \quad Token-level precision and recall against the gold answer, ensures strict factual correctness in question answering.
    
    \item \textbf{Normalized Levenshtein Distance} \quad Character-level string distance, indicating surface-level divergence (described by \citet{yujian2007normalized}).
    
    \item \textbf{Cosine Similarity} \quad Evaluates the semantic alignment, with robustness to minor changes.
    
    \item \textbf{LLM-as-a-Judge} \quad Gemini-2.0-Pro is prompted to score each prediction from 1 to 5 based on its semantic correctness, given the question and reference answer.

\end{itemize}

We exclude ROUGE, BLEU, and similar metrics because they are known to perform poorly on short factual answers, tend to over-penalize surface-level variations and are not well-aligned with human judgment in our setting \citep{blagec2022globalanalysismetricsused,chaganty2018pricedebiasingautomaticmetrics}.

\subsection{Human Evaluation Metrics}
\label{subapp:human_evaluation_metrics}

We conducted a human evaluation of the LLM-generated responses using three independent evaluators. Two of these evaluators were annotators familiar with credit agreements, while the third evaluator had no prior exposure to the documents. Each evaluator was requested to rate the LLM responses on a scale from 1 to 5, according to the criteria below. A total of 100 questions were sampled from multiple templates to ensure maximal diversity in the human evaluation task.

\begin{itemize}[leftmargin=*,nosep]
\item 1 \qquad response is irrelevant or does not match the actual answer;
\item 2 \qquad response is somewhat relevant but does not match the actual answer;
\item 3 \qquad response shows slight similarity but lacks accuracy;
\item 4 \qquad response is largely relevant and accurate;
\item 5 \qquad response is a perfect match to the actual answer.
\end{itemize}

\begin{figure*}[htbp]
    \centering
    \begin{subfigure}[b]{0.5\textwidth}
        \centering
        \includegraphics[width=\linewidth]{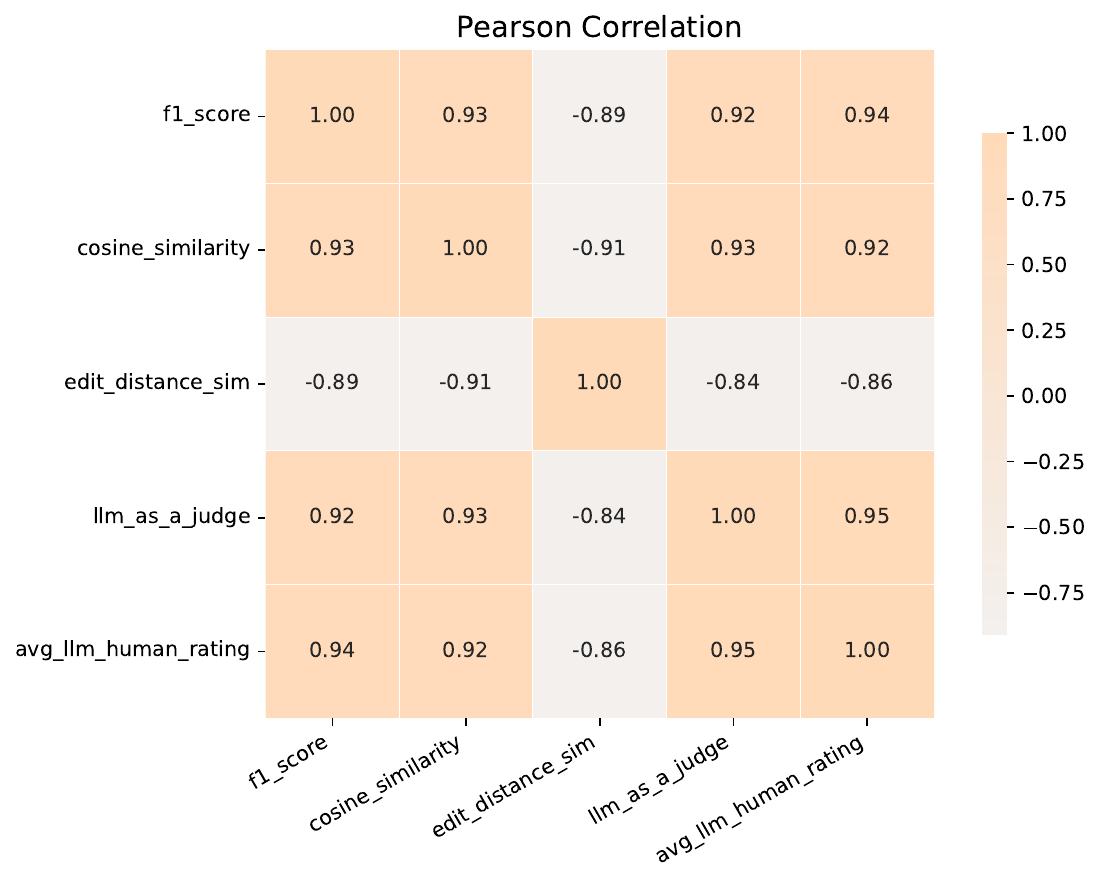}
      \caption{\textbf{Pearson correlation between various metrics.} F1 Score and Avg LLM-Human Rating (r = 0.941), and LLM-as-a-Judge and Avg LLM-Human Rating (r = 0.947) show the highest correlations in the matrix.}
      \label{fig:pearson_correlation}
    \end{subfigure}
    \hfill
    \begin{subfigure}[b]{0.47\textwidth}
        \centering
        \includegraphics[width=\linewidth]{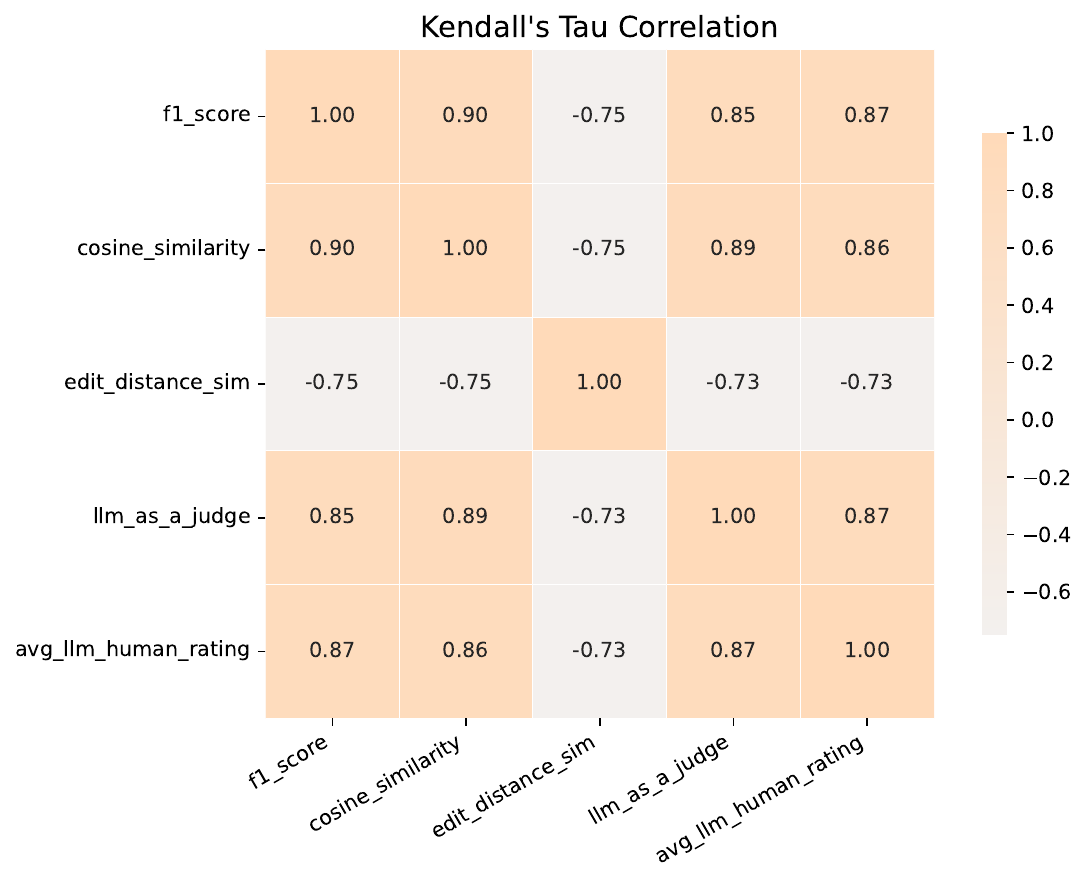}
        \caption{\textbf{Kendall-Tau correlation between various metrics.} The highest correlations are observed between F1 Score and Avg LLM-Human Rating (\( \tau \) = 0.870), and between LLM-as-a-Judge and Avg LLM-Human Rating (\( \tau \) = 0.867).}
        \label{fig:kendall_tau_correlation}
    \end{subfigure}
    \caption{Pearson and Kendall’s Tau Correlation Matrices.}
    \label{fig:metrics_correlations}
\end{figure*}

We assess the relationship between human evaluation scores and other metrics using both Pearson and Kendall’s Tau correlation coefficients (see Figure \ref{fig:metrics_correlations} and Table \ref{tab:metrics_correlation}). Pearson correlation captures linear relationships, while Kendall’s Tau assesses rank-based similarity, making them complementary for evaluating metric alignment. Across both measures, F1 Score and LLM-as-a-Judge consistently show the highest positive correlations with average human ratings—indicating not only strong agreement in scale but also in ranking. This indicates that both F1 Score and LLM-as-a-Judge closely align with human evaluations, supporting their use as strong performance metrics.

\paragraph{Metric prioritization and interpretation}
Based on the observed correlations with human judgments, we recommend interpreting LLM-as-a-Judge as the primary indicator of semantic correctness, particularly for higher-complexity questions where answers may be paraphrased, partially implicit, or expressed with variable surface forms. F1 score should be viewed as a complementary metric that captures extractive fidelity and coverage, and is most informative when exact grounding in the document is required. Discrepancies between the two metrics are therefore not treated as inconsistencies but as signals of different model behaviors (e.g., cautious but semantically accurate answers versus broader but partially grounded ones), as discussed in Section 3.3. For practical use, we recommend reporting both metrics jointly: LLM-as-a-Judge to assess answer correctness from a human-aligned perspective, and F1 to diagnose retrieval completeness and over-/under-generation. The remaining metrics (Edit Distance and Cosine Similarity) provide additional diagnostic insight into surface-level mismatch and semantic drift but are not intended to be used in isolation for model ranking.

\begin{table*}[h]
    \caption{Comparison of model performance across difficulty levels, including F1 Score, Normalized Edit Distance, Cosine Similarity, and LLM-as-a-Judge. $^{\ast}$ denotes the models evaluated on a smaller subset due to resources constraints.}
    \label{tab:full_results}

    \centering
    \resizebox{\textwidth}{!}{ 
    \rowcolors{4}{gray!20}{white} 
    \begin{tabular}{l|cccc|cccc|cccc}
        \hline
        & \multicolumn{4}{|c}{\textbf{Easy}} & \multicolumn{4}{|c}{\textbf{Medium}} & \multicolumn{4}{|c}{\textbf{Hard}} \\
        \cline{2-13}
        \textbf{Model} & F1 & Edit & Cosine & LLM-as- & F1 & Edit & Cosine & LLM-as- & F1 & Edit & Cosine & LLM-as- \\
        & Score $\uparrow$ & Dist. $\downarrow$ & Sim. $\uparrow$ & a-Judge $\uparrow$ & Score $\uparrow$ & Dist. $\downarrow$ & Sim. $\uparrow$ & a-Judge $\uparrow$ & Score $\uparrow$ & Dist. $\downarrow$ & Sim. $\uparrow$ & a-Judge $\uparrow$ \\
        \hline
        
        \multicolumn{13}{l}{\textbf{Proprietary LLMs}} \\
        Gemini-2.0-Flash$^{\ast}$ & \textbf{0.6748} & \textbf{0.2679} & \textbf{0.6650} & \textbf{4.0000} & 0.3294 & 0.6280 & 0.3121 & 2.4820 & 0.1172 & \textbf{0.7238} & 0.0608 & \textbf{2.2838} \\ 
        Gemini-2.0-Flash-Lite$^{\ast}$ & \underline{0.5908} & \underline{0.3763} & \underline{0.5829} & \underline{3.4172} & 0.3883 & \underline{0.5985} & \underline{0.3715} & \underline{2.6497} & \textbf{0.1943} & 0.7956 & \underline{0.1255} & 1.5135 \\
        Gemini-1.5-Pro$^{\ast}$ & 0.5410 & 0.3903 & 0.5331 & 3.2448 & \textbf{0.4577} & \textbf{0.5340} & \textbf{0.4415} & \textbf{2.8436} & \underline{0.1791} & 0.7786 & \textbf{0.1780} & \underline{1.6986} \\
        Gemini-1.5-Flash$^{\ast}$ & 0.5103 & 0.4285 & 0.5032 & 3.1207 & \underline{0.3892} & 0.6101 & 0.3657 & 2.5130 & 0.1400 & 0.8389 & \underline{0.1255} & 1.5411 \\
        GPT-4o$^{\ast}$ & 0.5272 & 0.4177 & 0.5213 & 3.2034 & 0.3783 & 0.6098 & 0.3690 & 2.5451 & 0.1178 & \underline{0.7738} & 0.1144 & 1.4795 \\
        GPT-4o-Mini$^{\ast}$ & 0.4723 & 0.4614 & 0.4654 & 2.9690 & 0.2732 & 0.6756 & 0.2740 & 2.0681 & 0.0274 & 0.8294 & 0.0336 & 1.1370 \\
        
        \hline
        
        \multicolumn{13}{l}{\textbf{Open-source LLMs}} \\
        DeepSeek-R1$^{\ast}$ & \textbf{0.6269} & \textbf{0.3510} & \textbf{0.6208} & \textbf{3.6965} & 0.3293 & 0.5191 & 0.3179 & 2.6354 & \underline{0.2609} & 0.7733 & \underline{0.1451} & 2.0736 \\
        DeepSeek-V3 & \underline{0.6218} & 0.3722 & \underline{0.6077} & 3.5668 & \textbf{0.4796} & \underline{0.4912} & \underline{0.4678} & \underline{3.0210} & 0.0856 & 0.7975 & 0.0737 & 1.4795 \\
        GPT OSS 120B & 0.5885 & 0.4187 & 0.5776 & \underline{3.5839} & \underline{0.4684} & \textbf{0.4765} & \textbf{0.4740} & \textbf{3.1661} & 0.1136 & \textbf{0.7067} & 0.1202 & 1.4888 \\
        Llama-4-Maverick-Instruct$^{\ast}$ & 0.5552 & \underline{0.3585} & 0.5494 & 3.4482 & 0.3848 & 0.5606 & 0.3669 & 2.6976 & 0.2031 & 0.7231 & 0.1172 & \textbf{3.1351} \\
        Llama-4-Scout-Instruct$^{\ast}$ & 0.4993 & 0.4114 & 0.4920 & 3.2138 & 0.2215 & 0.6765 & 0.2099 & 1.8772 & 0.0472 & 0.8225 & 0.0303 & 1.5270 \\
        Llama-3.1-Instruct-Turbo$^{\ast}$ & 0.5472 & 0.3901 & 0.5389 & 3.4793 & 0.4204 & 0.5489 & 0.4018 & 2.6377 & \textbf{0.3651} & \underline{0.7197} & \textbf{0.2849} & \underline{2.7297} \\
        Llama-3.1-Instruct & 0.4955 & 0.4707 & 0.4847 & 3.1373 & 0.4017 & 0.5481 & 0.3917 & 2.6722 & 0.1639 & 0.7667 & 0.1106 & 2.3425 \\
        Llama-3.1-Instruct & 0.3891 & 0.5617 & 0.3793 & 2.6996 & 0.2962 & 0.6255 & 0.2856 & 2.2332 & 0.1169 & 0.7980 & 0.0915 & 1.4315 \\
        Qwen3-Instruct & 0.5243 & 0.4243 & 0.5133 & 3.2693 & 0.2760 & 0.5795 & 0.2670 & 2.4460 & 0.1240 & 0.7473 & 0.1129 & 1.9578 \\
        Qwen2-Instruct & 0.3886 & 0.6369 & 0.3710 & 3.3884 & 0.3100 & 0.7031 & 0.2986 & 2.9078 & 0.0762 & 0.8200 & 0.0690 & 1.3288 \\
        \hline
 \end{tabular}
}
\end{table*}

\begin{table*}[htbp]

\caption{Pearson and Kendall’s Tau Correlation Tables.}
\label{tab:metrics_correlation}
\centering

\resizebox{\textwidth}{!}{
\begin{tabular}{lccccc}
\toprule
\multicolumn{6}{c}{\textbf{Pearson Correlation Matrix}} \\
\midrule
 & \textbf{F1 Score} & \textbf{Cosine Similarity} & \textbf{Edit Distance Sim} & \textbf{LLM-as-a-Judge} & \textbf{Avg LLM-Human Rating} \\
\textbf{F1 Score} & 1.000 & 0.933 & -0.894 & 0.922 & 0.941 \\
\textbf{Cosine Similarity} & 0.933 & 1.000 & -0.910 & 0.935 & 0.922 \\
\textbf{Edit Distance Sim} & -0.894 & -0.910 & 1.000 & -0.844 & -0.862 \\
\textbf{LLM-as-a-Judge} & 0.922 & 0.935 & -0.844 & 1.000 & 0.947 \\
\textbf{Avg LLM-Human Rating} & 0.941 & 0.922 & -0.862 & 0.947 & 1.000 \\
\midrule
\midrule
\multicolumn{6}{c}{\textbf{Kendall’s Tau Correlation Matrix}} \\
\midrule
 & \textbf{F1 Score} & \textbf{Cosine Similarity} & \textbf{Edit Distance Sim} & \textbf{LLM-as-a-Judge} & \textbf{Avg LLM-Human Rating} \\
\textbf{F1 Score} & 1.000 & 0.902 & -0.745 & 0.854 & 0.870 \\
\textbf{Cosine Similarity} & 0.902 & 1.000 & -0.746 & 0.894 & 0.858 \\
\textbf{Edit Distance Sim} & -0.745 & -0.746 & 1.000 & -0.734 & -0.733 \\
\textbf{LLM-as-a-Judge} & 0.854 & 0.894 & -0.734 & 1.000 & 0.867 \\
\textbf{Avg LLM-Human Rating} & 0.870 & 0.858 & -0.733 & 0.867 & 1.000 \\
\bottomrule
\end{tabular}%
}
\end{table*}

\clearpage

\subsection{Cost-Constrained Models}
\label{subapp:cost_constrained_models}

For the cost-constrained models, we used stratified sampling over our question templates. Because each template corresponds to a unique point in the (H, P, \#SO) space, sampling proportionally over templates preserves the difficulty distribution, the multi-hop vs single-hop ratio, and the relative frequency of set-operation queries. Using DeepSeek-V3, the strongest open-weight model evaluated on the full dataset, we have verified that performance on the sampled subset correlates strongly with full-set performance: 0.87 for F-score, 0.84 for edit distance, and 0.88 for cosine similarity.

\subsection{Chunk-Based Evaluation}
\label{subapp:chunk_based_evaluation}

As seen in Table \ref{tab:no_chunking_results}, the effect of chunking becomes noticeable as the question complexity grows. The performance increase is driven mainly by a reduction in abstentions, with models more often returning explicit answers instead of \texttt{``Not Found''}. However, even for medium questions nearly half is still unanswerable without chunking: it does not reflect improved reasoning or extraction quality, which means chunking does not affect our error analysis or the conclusions drawn in the paper.

\begin{table}[h]
\caption{Evaluation of Gemini-2.0-Flash with chunking (as in the main paper) and without chunking. The context length of this model is large enough to incorporate an entire credit agreement in one prompt.}
\label{tab:no_chunking_results}
\centering
\resizebox{\linewidth}{!}{
\begin{tabular}{lcccc}
\hline
\textbf{Strategy} & \textbf{F1 $\uparrow$} & \textbf{Edit $\downarrow$} & \textbf{Cos $\uparrow$} & \textbf{LLM-J $\uparrow$} \\
\toprule
\multicolumn{5}{c}{\textit{Easy}} \\
With & 0.6748 & 0.2679 & 0.6650 & 4.0000 \\
Without & 0.7105 & 0.2987 & 0.6949 & 3.9269 \\
\midrule
\multicolumn{5}{c}{\textit{Medium}} \\
With & 0.3294 & 0.6280 & 0.3121 & 2.4820 \\
Without & 0.5558 & 0.4051 & 0.5452 & 3.4344 \\
\midrule
\multicolumn{5}{c}{\textit{Hard}} \\
With & 0.1172 & 0.7238 & 0.0608 & 2.2838 \\
Without & 0.2565 & 0.6310 & 0.2504 & 2.1088 \\
\bottomrule
\end{tabular}
}
\end{table}

\subsection{Single Question Evaluation}
\label{subapp:single_question_evaluation}

We note that the absolute metric values in Table \ref{tab:single_question_eval} are not representative of task complexity, as a sampling of 20 questions per complexity level was required due to the high cost of single-question evaluation for long-context documents. Examining relative differences, the single-question evaluation provides marginal gains for easy and medium questions, but not for hard ones. These variations are minor and leave the overall performance patterns across complexity levels unchanged.

\begin{table}[h]
\caption{Evaluation of Gemini-2.0-Flash in batches of 50 questions (as in the main paper) and with a simple questions.}
\label{tab:single_question_eval}
\centering
\resizebox{\linewidth}{!}{
\begin{tabular}{lcccc}
\toprule
\textbf{Difficulty} & \textbf{F1 $\uparrow$} & \textbf{Edit $\downarrow$} & \textbf{Cos $\uparrow$} & \textbf{LLM-J $\uparrow$} \\
\hline
\multicolumn{5}{c}{\textit{Easy}} \\
Batch & 0.5277 & 0.3773 & 0.5249 & 3.7500 \\
Single & 0.5719 & 0.3573 & 0.5688 & 3.8000 \\
\midrule
\multicolumn{5}{c}{\textit{Medium}} \\
Batch & 0.3980 & 0.4220 & 0.3958 & 3.2040 \\
Single & 0.4451 & 0.3695 & 0.4427 & 3.5510 \\
\midrule
\multicolumn{5}{c}{\textit{Hard}} \\
Batch & 0.5333 & 0.4223 & 0.5333 & 3.1333 \\
Single & 0.4971 & 0.4367 & 0.4938 & 3.0666 \\
\bottomrule
\end{tabular}
}
\end{table}

\subsection{Human Performance Evaluation}
\label{subapp:human_performance_eval}

To establish a reference point for model comparison, we conducted a human evaluation on a balanced subset of 100 questions (20 per difficulty level). Three independent human evaluators participated in this study: one annotator who was fully familiar with the documents, and two participants with general financial knowledge but no prior exposure to the documents. Their responses were normalized and compared against ground truth using three similarity metrics: word level F1, edit distance and cosine similarity. Unlike model performance, which typically declines as task complexity increases, human performance remained relatively stable across levels, as seen in Table \ref{tab:human-performance}. This suggests that humans effectively leverage contextual reasoning and background knowledge to maintain accuracy even in multi-hop or compositional settings, capabilities that current models still struggle with.

\begin{table}[h]
\centering
\caption{Mean human performance across question levels.}
\label{tab:human-performance}
\begin{tabular}{lccc}
\toprule
\textbf{Level} & \textbf{F1} & \textbf{Edit Dist.} & \textbf{Cosine Sim.} \\
\midrule
1 & 0.846 & 0.142 & 0.834 \\
2 & 0.778 & 0.264 & 0.750 \\
3 & 0.780 & 0.235 & 0.756 \\
4 & 0.803 & 0.244 & 0.795 \\
5 & 0.935 & 0.101 & 0.928 \\
\bottomrule
\end{tabular}
\end{table}

\begin{table*}[htbp]
    \caption{Performance comparison across four retrieval settings -- Full (entire credit agreement), Oracle (gold snippet only), RAG and dynamic RAG (retriever-selected passages) -- for each question complexity group.}
    \label{tab:results_with_settings}
    
    \centering
    \resizebox{\textwidth}{!}{ 
    \rowcolors{4}{gray!20}{white} 
    \begin{tabular}{c|cccc|cccc|cccc}
        \hline
        & \multicolumn{4}{|c}{\textbf{Easy}} & \multicolumn{4}{|c}{\textbf{Medium}} & \multicolumn{4}{|c}{\textbf{Hard}} \\
        \cline{2-13}
        \textbf{Setting} & F1 & Edit & Cosine & LLM-as- & F1 & Edit & Cosine & LLM-as- & F1 & Edit & Cosine & LLM-as- \\
        & Score $\uparrow$ & Dist. $\downarrow$ & Sim. $\uparrow$ & a-Judge $\uparrow$ & Score $\uparrow$ & Dist. $\downarrow$ & Sim. $\uparrow$ & a-Judge $\uparrow$ & Score $\uparrow$ & Dist. $\downarrow$ & Sim. $\uparrow$ & a-Judge $\uparrow$ \\
        \hline
        
        Full & 0.6747 & 0.2678 & 0.6650 & \textbf{4.000} & \textbf{0.3294} & 0.6279 & 0.3121 & 2.4820 & \textbf{0.1172} & \textbf{0.7237} & 0.0607 & \textbf{2.2837} \\
        Oracle & \textbf{0.7134} & \textbf{0.2588} & \textbf{0.7021} & 3.9862 & \textbf{0.3294} & \textbf{0.5986} & \textbf{0.3201} & \textbf{2.6497} & 0.1127 & 0.8066 & \textbf{0.0735} & 1.9189 \\
        RAG & 0.1579 & 0.7364 & 0.1546 & 1.6931 & 0.0695 & 0.8070 & 0.0651 & 1.4131 & 0.0190 & 0.8449 & 0.0107 & 1.2027 \\
        D-RAG & 0.1671 & 0.7304 & 0.1631 & 1.7185 & 0.1118 & 0.7693 & 0.1087 & 1.5566 & 0.0107 & 0.8431 & 0.0069 & 1.1088 \\
        
        \hline
    \end{tabular}
    }
\end{table*}

\subsection{Comparison with Oracle and RAG}
\label{subapp:rag_oracle_results}

We compare four retrieval settings for answering questions from credit agreements: (1) \textit{Full}, where the entire agreement is provided to the model, (2) \textit{Oracle}, which uses the pieces of text containing the answer; (3) \textit{RAG} (Retrieval-Augmented Generation), which selects top passages using a retriever; and (4) \textit{Dynamic RAG}, which iteratively decomposes the question into sub-queries, retrieves relevant passages across multiple steps, and aggregates the retrieved evidence before generating the final answer. In Table \ref{tab:results_with_settings}, we observe that RAG, both simple and dynamic, performs drastically worse than the other two settings across all difficulty levels, which highlights the incompatibility of standard retrieval techniques with dense, cross-referenced financial documents, often requiring the aggregation of logically related but spatially distant elements.

Interestingly, the comparison between Oracle and Full Document settings reveals a non-trivial pattern. For easy look-up and for medium questions, the Oracle setting performs better, as expected, since such questions benefit from isolating the minimal, most relevant snippet. However, for hard questions, the Full Document consistently outperforms the Oracle setting. This suggests that isolating the ``correct'' snippet removes necessary contextual cues that models need for reasoning, \textbf{\textit{which is detailed in the Implicit Information Gaps paragraph of our Error Analysis}} (Section \ref{subsec:error_analysis}). These findings reinforce the value of credit agreements as a testbed for long-context LLMs, and demonstrate how our framework's question complexity help uncover the precise scenarios where models struggle, even when given the exact ground-truth passage (see Section \ref{sec:analysis} and \ref{app:error_analysis} for the detailed analysis).

\section{Error Analysis}
\label{app:error_analysis}

\begin{figure*}[htbp]
    \centering
    \begin{subfigure}[b]{0.9\textwidth}
        \centering
        \includegraphics[width=\textwidth]{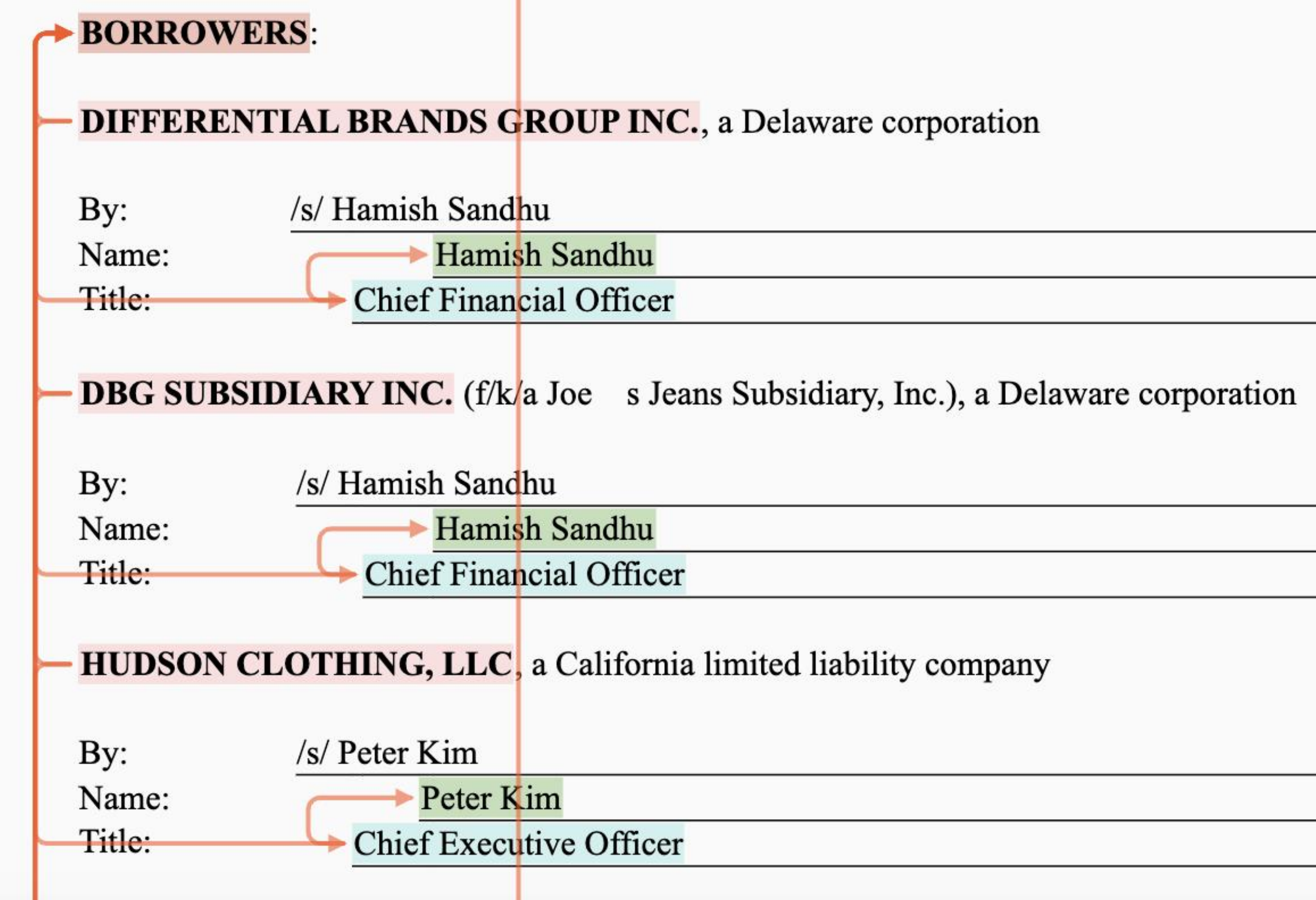}
        \caption{Multiple entities and roles are listed in close proximity, making it difficult for the model to associate roles accurately.}
        \label{fig:dense_signature_panel}
    \end{subfigure}
    
    \begin{subfigure}[b]{0.3\textwidth}
        \centering
        \includegraphics[width=\textwidth]{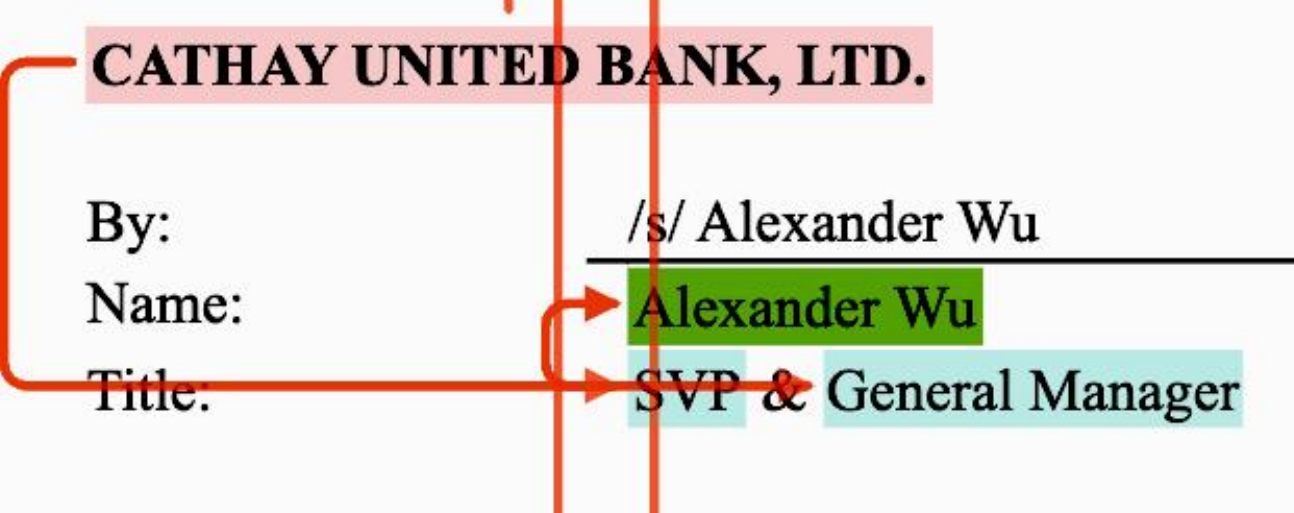}
        \caption{Alexander Wu}
        \label{fig:alexander}
    \end{subfigure}
    \hfill
    \begin{subfigure}[b]{0.3\textwidth}
        \centering
        \includegraphics[width=\textwidth]{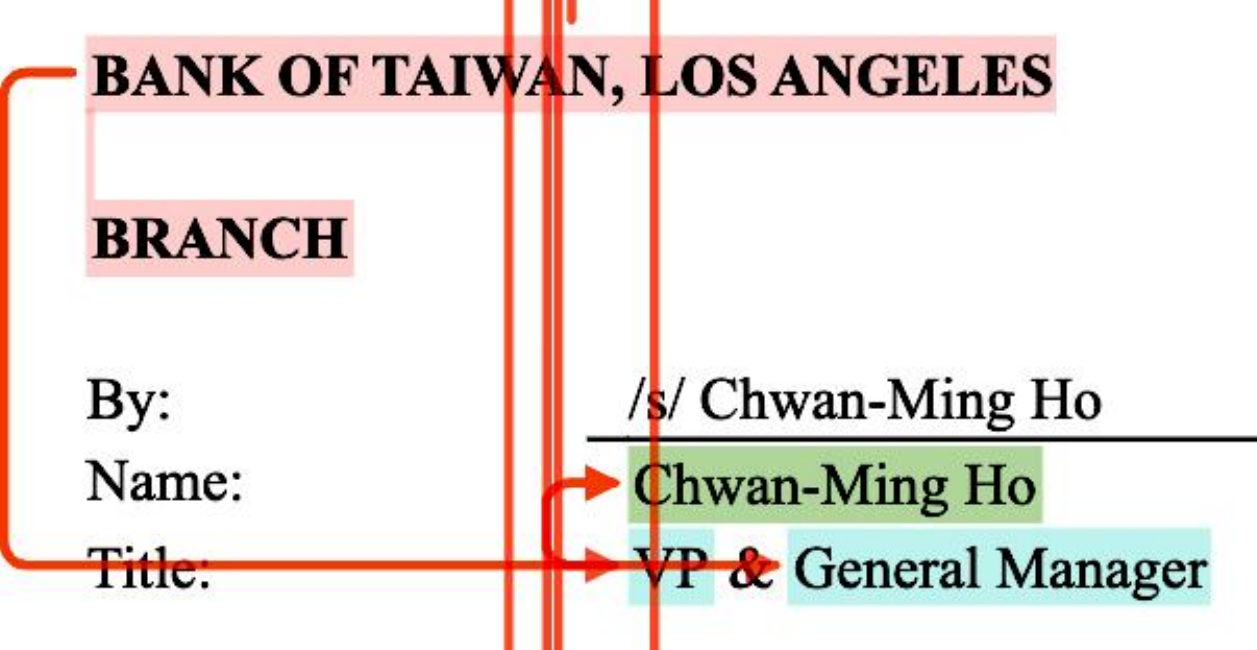}
        \caption{Chwan Ming Ho}
        \label{fig:chwan}
    \end{subfigure}
    \hfill
    \begin{subfigure}[b]{0.3\textwidth}
        \centering
        \includegraphics[width=\textwidth]{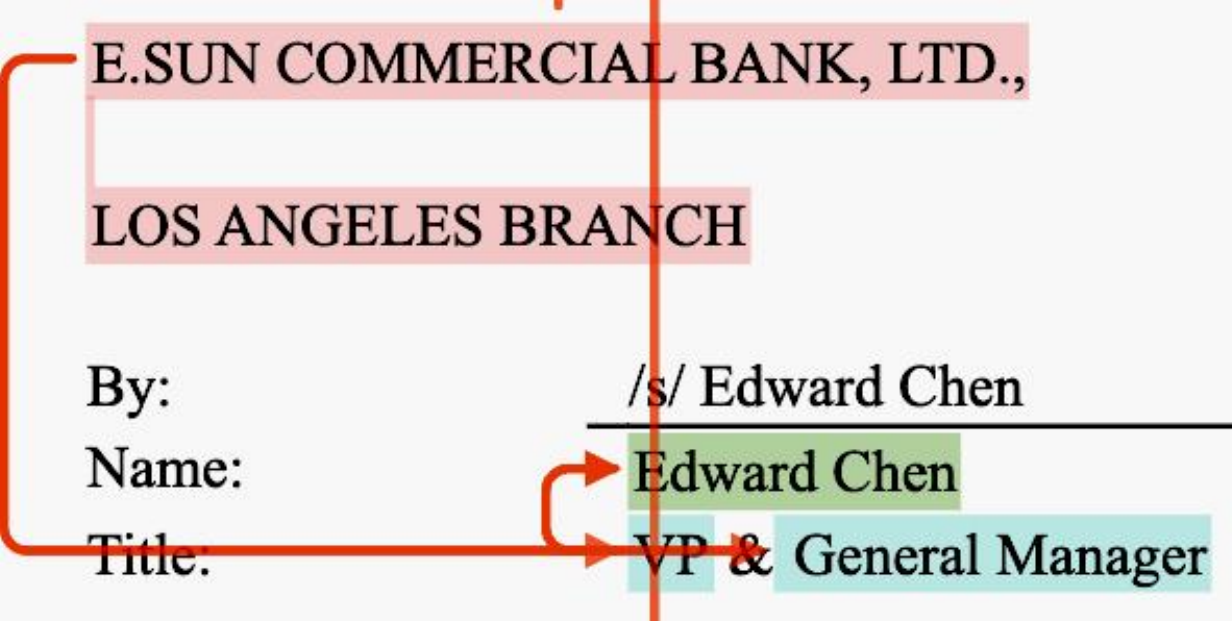}
        \caption{Edward Chen}
        \label{fig:edward}
    \end{subfigure}
    
    \caption{(a) In cases involving multiple signatories, the LLM cannot infer implicit information like Person Positions. (b), (c) and (d) illustrate the LLM's inability to deal with multiple set operations.}
    \label{fig:error_analysis_example}
\end{figure*}

\begin{figure*}[htbp]
    \centering
    \includegraphics[width=\linewidth]{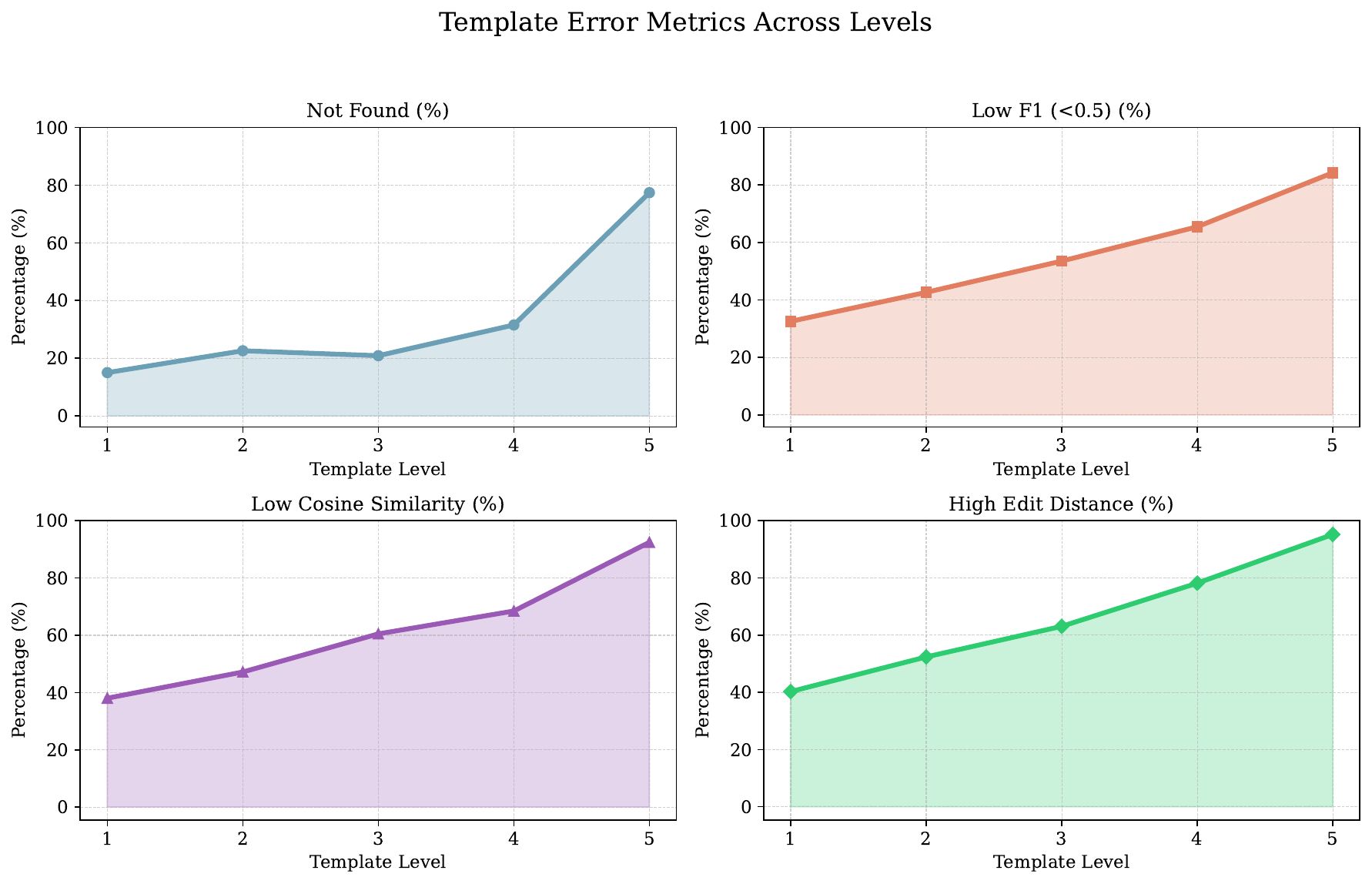}
    \caption{\textbf{Trends in Error Metrics Across Template Levels.} The plots illustrate a clear upward trend across all error metrics as template complexity increases. This includes the percentage of LLM responses marked as "Not Found", as well as those exhibiting low F1 scores, low cosine similarity, and high edit distance -- indicating a decline in accuracy with higher-level templates.}
    \label{fig:template_errors}
\end{figure*}

\begin{figure*}[htbp]
  \centering

  \begin{subfigure}[b]{0.48\textwidth}
    \centering
    \includegraphics[width=\textwidth]{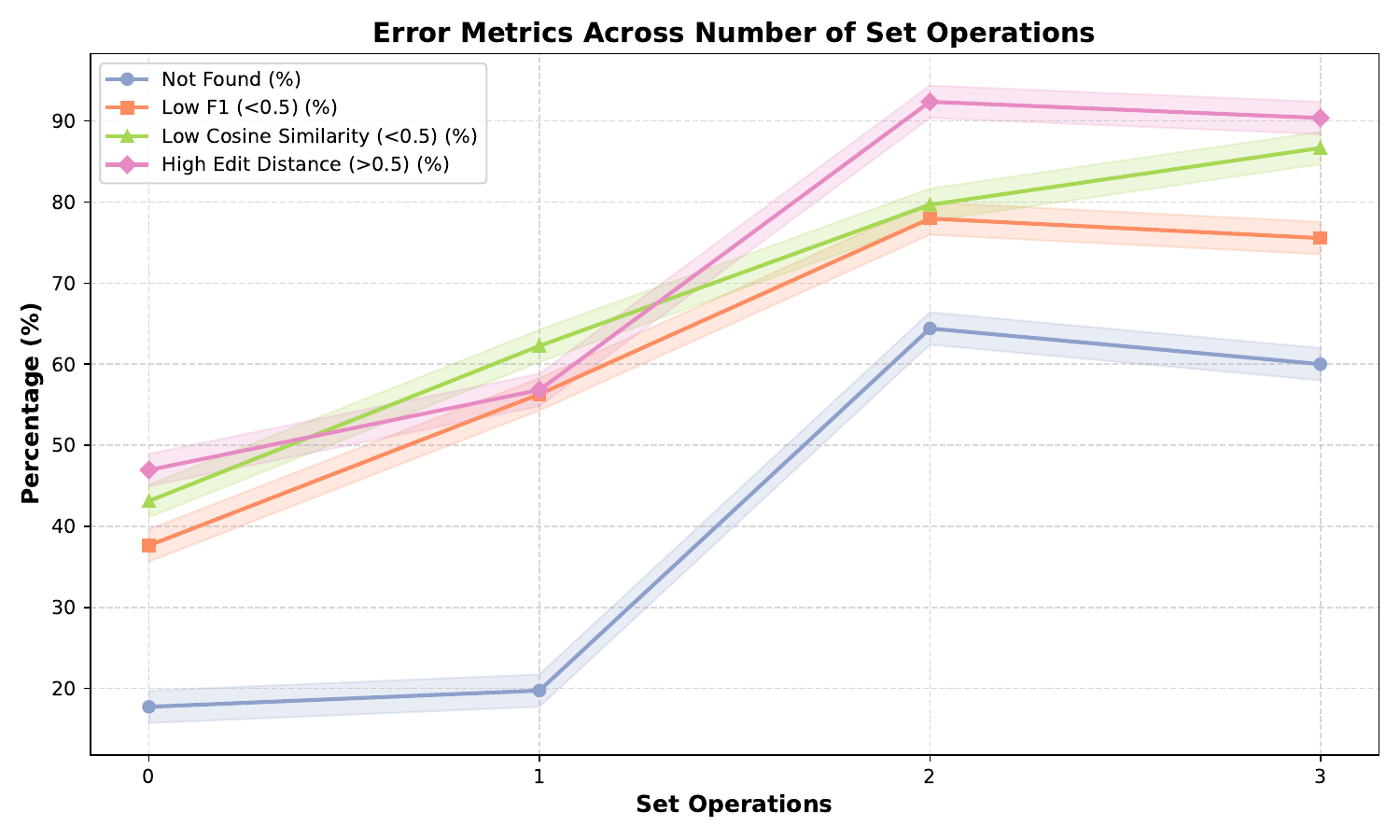}
    \caption{Error metrics across set operations.}
    \label{fig:set_ops_plot}
  \end{subfigure}%
  \hfill
  \begin{subfigure}[b]{0.48\textwidth}
    \centering
    \includegraphics[width=\textwidth]{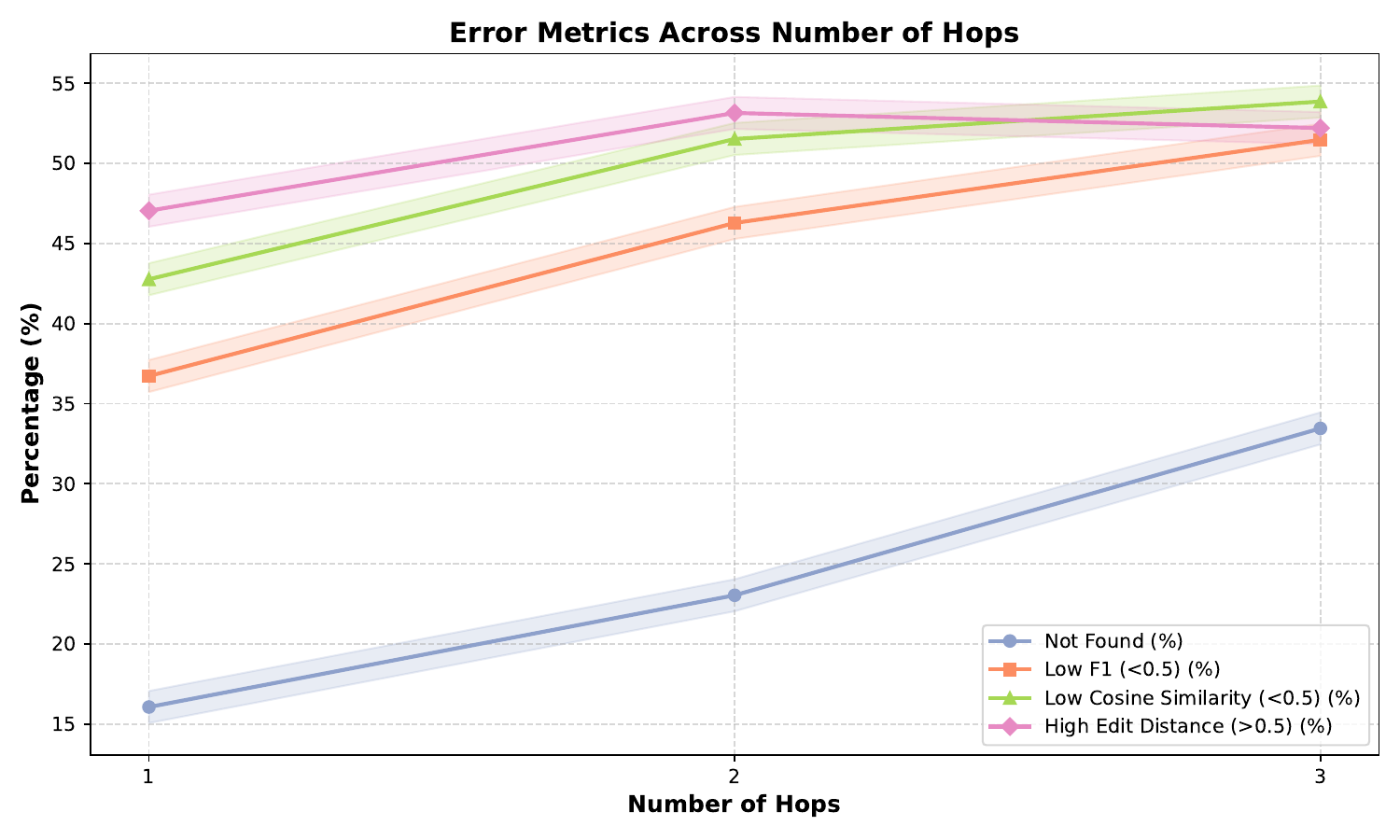}
    \caption{Error metrics across number of hops.}
    \label{fig:num_hops_plot}
  \end{subfigure}
  \caption{\textbf{Comparison of error metrics for Deepseek-V3.} (a) Trends by Set Operations. (b) Trends by Number of Hops. With increase in number of hops or set operations, the percentage distribution of each error type increases.}
  \label{fig:error_metrics_comparison_by_dimension}
\end{figure*}

Detailed examples of errors from each of the aforementioned categories are shown in Table \ref{tab:error_analysis_examples} and Figure \ref{fig:error_analysis_example}. Aggregate trends across error types and complexity are shown in Figures \ref{fig:template_errors} and \ref{fig:error_metrics_comparison_by_dimension}.

\begin{table*}[htbp]
    \centering
    \small 
    \caption{Error analysis with examples.}

    \renewcommand{\arraystretch}{1.2} 
    \setlength{\tabcolsep}{5pt}      
    
    \begin{tabular}{>{\bfseries}p{3.5cm} p{11cm}} 
    \toprule
    Question & \textit{What type of location is Dubai, UAE?} \\
    Expected Answer & \textit{Zonal Office} \\
    LLM Answer & \textit{City} \\
    Error Explanation & \textbf{Misinterpretation of Semantics.} The LLM misinterprets the semantic intent of the question. While \textit{City} is factually accurate, the question requires identifying the Location Type of "Dubai, UAE" in the context of the credit agreement. The expected answer, \textit{zonal office,} reflects a more precise understanding. \\
    \midrule
    Question & \textit{In what organizations does Hamish Sandhu work?} \\
    Expected Answer & \textit{DBG Holdings Subsidiary Inc, Differential Brands Group Inc} \\
    LLM Answer & \textit{RG Parent LLC} \\
    Error Explanation & \textbf{Implicit Information Gaps.} The LLM fails to infer implicit information due to structural complexity. Human readers can associate individuals with their organizations based on layout and context. The LLM incorrectly attributes Hamish Sandhu to the wrong organization. \\
    \midrule
    Question & \textit{What is the position held by Alexander Wu but not by Chwan Ming Ho or Edward Chen?} \\
    Expected Answer & \textit{SVP} \\
    LLM Answer & \textit{SVP and General Manager} \\
    Error Explanation & \textbf{Set Operation Failures.} The LLM struggles with multiple set operations. Alexander Wu is SVP and General Manager, while Chwan-Ming Ho and Edward Chen are both VPs and General Managers. The extra "General Manager" indicates a failure in subtracting shared titles. \\
    \midrule
    Question & \textit{What is the role of Celanese Chemicals, Inc. in the agreement?} \\
    Expected Answer & \textit{Subsidiary} \\
    LLM Answer & \textit{Not found} \\
    Error Explanation & \textbf{Long-Context Retrieval Limitations.} The model fails because "subsidiary" is not explicitly stated in a natural sentence. Instead, Celanese Chemicals, Inc. appears under headings or clauses indicating subsidiaries, requiring structural and hierarchical reasoning. \\
    \bottomrule
    \end{tabular}

    \label{tab:error_analysis_examples}
\end{table*}

\clearpage

\section{Ablation Study on Medical Documents}
\label{app:ablation_study_medical}

\begin{figure}[h]
  \centering
  \includegraphics[width=\linewidth]{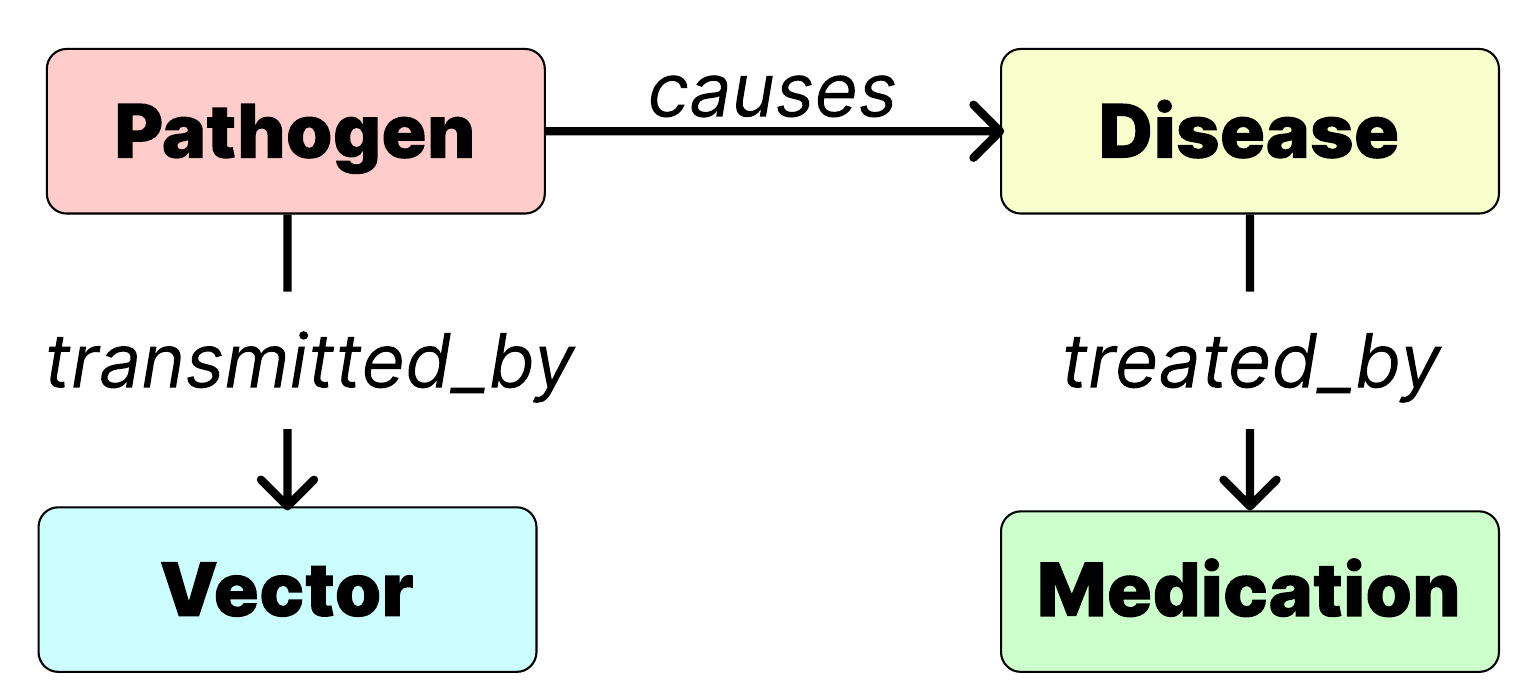}
  \caption{Annotation schema of our ablation study.}
  \label{fig:ablation_schema}
\end{figure}

To illustrate the generalizability and applicability of our framework to other domain, we conducted a minor ablation study on the publicly available\footnote{https://clinicalinfo.hiv.gov/en/guidelines/hiv-clinical-guidelines-adult-and-adolescent-opportunistic-infections/whats-new} "Guidelines for the Prevention and Treatment of Opportunistic Infections in Adults and Adolescents With HIV". We extracted the introductory part, the first 10 descriptions of diseases, and the conclusive part, which results in a 225-page document. We annotated the document according to the schema in Figure \ref{fig:ablation_schema}, which allowed us to extract questions of various complexity levels over the same 3 dimensions as in the main study (see Tables \ref{tab:ablation_templatex_L1}-\ref{tab:ablation_templatex_L3} for the full list of question templates).

We ran our experiments with the best-performing open-weight model, DeepSeek-V3. In Table \ref{tab:ablation_results}, we observe a clear performance decrease as the question complexity increases, which illustrates that our framework is generalizable and applicable to other domains.

\begin{table}[h]
\caption{Evaluation of DeepSeek-V3 on the ablation study dataset.}
\label{tab:ablation_results}
\centering
\resizebox{\linewidth}{!}{
\begin{tabular}{lcccc}
\hline
\textbf{Difficulty} & \textbf{F1 $\uparrow$} & \textbf{Edit $\downarrow$} & \textbf{Cos $\uparrow$} & \textbf{LLM-J $\uparrow$} \\
\hline
Easy    & 0.2585 & 0.6169 & 0.2522 & 2.2249 \\
Medium  & 0.0753 & 0.7512 & 0.0662 & 1.5587 \\
Hard    & 0.0    & 0.8417 & 0.0    & 1.0681 \\
\hline
\end{tabular}
}
\end{table}

\begin{table*}[htbp]
    \caption{Ablation 1-hop Question Templates}
    \centering
    \resizebox{\textwidth}{!}{
    \begin{tabular}{p{0.85\textwidth}p{0.05\textwidth}p{0.05\textwidth}p{0.05\textwidth}}
    \toprule
    \textbf{Template} & \textbf{P} & \textbf{H} & \textbf{\#SO} \\
    \midrule
    What pathogen is transmitted by [Vector]? [if one] & 0 & 1 & 0 \\
    What vector transmits [Pathogen]? [if one] & 0 & 1 & 0 \\
    What pathogen causes [Disease]? [if one] & 0 & 1 & 0 \\
    What disease is caused by [Pathogen]? [if one] & 0 & 1 & 0 \\
    What disease is treated by [Medication]? [if one] & 0 & 1 & 0 \\
    What medication treats [Disease]? [if one] & 0 & 1 & 0 \\
    \midrule
    What pathogens are transmitted by [Vector]? & 1 & 1 & 0 \\
    What vectors transmit [Pathogen]? & 1 & 1 & 0 \\
    What pathogens cause [Disease]? & 1 & 1 & 0 \\
    What diseases are caused by [Pathogen]? & 1 & 1 & 0 \\
    What diseases are treated by [Medication]? & 1 & 1 & 0 \\
    What medications treat [Disease]? & 1 & 1 & 0 \\
    \midrule
    What pathogen is transmitted by [Vector 1] and [Vector 2]? [if one] & 0 & 1 & 1 \\
    What vector transmits [Pathogen 1] and [Pathogen 2]? [if one] & 0 & 1 & 1 \\
    What pathogen causes [Disease 1] and [Disease 2]? [if one] & 0 & 1 & 1 \\
    What disease is caused by [Pathogen 1] and [Pathogen 2]? [if one] & 0 & 1 & 1 \\
    What disease is treated by [Medication 1] and [Medication 2]? [if one] & 0 & 1 & 1 \\
    What medication treats [Disease 1] and [Disease 2]? [if one] & 0 & 1 & 1 \\
    \midrule
    What pathogens are transmitted by [Vector 1] and [Vector 2]? & 1 & 1 & 1 \\
    What vectors transmit [Pathogen 1] and [Pathogen 2]? & 1 & 1 & 1 \\
    What pathogens cause [Disease 1] and [Disease 2]? & 1 & 1 & 1 \\
    What diseases are caused by [Pathogen 1] and [Pathogen 2]? & 1 & 1 & 1 \\
    What diseases are treated by [Medication 1] and [Medication 2]? & 1 & 1 & 1 \\
    What medications treat [Disease 1] and [Disease 2]? & 1 & 1 & 1 \\
    \midrule
    What pathogen is transmitted by [Vector 1] but not [Vector 2]? [if one] & 0 & 1 & 2 \\
    What vector transmits [Pathogen 1] but not [Pathogen 2]? [if one] & 0 & 1 & 2 \\
    What pathogen causes [Disease 1] but not [Disease 2]? [if one] & 0 & 1 & 2 \\
    What disease is caused by [Pathogen 1] but not [Pathogen 2]? [if one] & 0 & 1 & 2 \\
    What disease is treated by [Medication 1] but not [Medication 2]? [if one] & 0 & 1 & 2 \\
    What medication treats [Disease 1] but not [Disease 2]? [if one] & 0 & 1 & 2 \\
    \midrule
    What pathogens are transmitted by [Vector 1] but not [Vector 2]? & 1 & 1 & 2 \\
    What vectors transmit [Pathogen 1] but not [Pathogen 2]? & 1 & 1 & 2 \\
    What pathogens cause [Disease 1] but not [Disease 2]? & 1 & 1 & 2 \\
    What diseases are caused by [Pathogen 1] but not [Pathogen 2]? & 1 & 1 & 2 \\
    What diseases are treated by [Medication 1] but not [Medication 2]? & 1 & 1 & 2 \\
    What medications treat [Disease 1] but not [Disease 2]? & 1 & 1 & 2 \\
    \bottomrule
    \end{tabular}
    }
    \label{tab:ablation_templatex_L1}
\end{table*}

\begin{table*}[htbp]
    \caption{Ablation 2-hop Question Templates}
    \centering
    \resizebox{\textwidth}{!}{
    \begin{tabular}{p{0.85\textwidth}p{0.05\textwidth}p{0.05\textwidth}p{0.05\textwidth}}
    \toprule
    \textbf{Template} & \textbf{P} & \textbf{H} & \textbf{\#SO} \\
    \midrule
    What vector transmits a pathogen which causes [Disease]? [if one] & 0 & 2 & 0 \\
    What disease is caused by a pathogen which is transmitted by [Vector]? [if one] & 0 & 2 & 0 \\
    What medication treats a disease which is caused by [Pathogen]? [if one] & 0 & 2 & 0 \\
    What pathogen causes a disease which is treated by [Medication]? [if one] & 0 & 2 & 0 \\
    \midrule
    What vectors transmit a pathogen which causes [Disease]? & 1 & 2 & 0 \\
    What diseases are caused by a pathogen which is transmitted by [Vector]? & 1 & 2 & 0 \\
    What medications treat a disease which is caused by [Pathogen]? & 1 & 2 & 0 \\
    What pathogens cause a disease which is treated by [Medication]? & 1 & 2 & 0 \\
    \midrule
    What vector transmits a pathogen which causes [Disease 1] and [Disease 2]? [if one] & 0 & 2 & 1 \\
    What disease is caused by a pathogen which is transmitted by [Vector 1] and [Vector 2]? [if one] & 0 & 2 & 1 \\
    What medication treats a disease which is caused by [Pathogen 1] and [Pathogen 2]? [if one] & 0 & 2 & 1 \\
    What pathogen causes a disease which is treated by [Medication 1] and [Medication 2]? [if one] & 0 & 2 & 1 \\
    \midrule
    What vectors transmit a pathogen which causes [Disease 1] and [Disease 2]? & 1 & 2 & 1 \\
    What diseases are caused by a pathogen which is transmitted by [Vector 1] and [Vector 2]? & 1 & 2 & 1 \\
    What medications treat a disease which is caused by [Pathogen 1] and [Pathogen 2]? & 1 & 2 & 1 \\
    What pathogens cause a disease which is treated by [Medication 1] and [Medication 2]? & 1 & 2 & 1 \\
    \midrule
    What vector transmits a pathogen which causes [Disease 1] but not [Disease 2]? [if one] & 0 & 2 & 2 \\
    What disease is caused by a pathogen which is transmitted by [Vector 1] but not [Vector 2]? [if one] & 0 & 2 & 2 \\
    What medication treats a disease which is caused by [Pathogen 1] but not [Pathogen 2]? [if one] & 0 & 2 & 2 \\
    What pathogen causes a disease which is treated by [Medication 1] but not [Medication 2]? [if one] & 0 & 2 & 2 \\
    \midrule
    What vectors transmit a pathogen which causes [Disease 1] but not [Disease 2]? & 1 & 2 & 2 \\
    What diseases are caused by a pathogen which is transmitted by [Vector 1] but not [Vector 2]? & 1 & 2 & 2 \\
    What medications treat a disease which is caused by [Pathogen 1] but not [Pathogen 2]? & 1 & 2 & 2 \\
    What pathogens cause a disease which is treated by [Medication 1] but not [Medication 2]? & 1 & 2 & 2 \\
    \bottomrule
    \end{tabular}
    }
    \label{tab:ablation_templatex_L2}
\end{table*}

\begin{table*}[htbp]
    \caption{Ablation 3-hop Question Templates}
    \centering
    \resizebox{\textwidth}{!}{
    \begin{tabular}{p{0.85\textwidth}p{0.05\textwidth}p{0.05\textwidth}p{0.05\textwidth}}
    \toprule
    \textbf{Template} & \textbf{P} & \textbf{H} & \textbf{\#SO} \\
    \midrule
    What vector transmits a pathogen which causes a disease which is treated by [Medication]? [if one] & 0 & 3 & 0 \\
    What medication treats a disease which is caused by a pathogen which is transmitted by [Vector]? [if one] & 0 & 3 & 0 \\
    \midrule
    What vectors transmit a pathogen which causes a disease which is treated by [Medication]? & 1 & 3 & 0 \\
    What medications treat a disease which is caused by a pathogen which is transmitted by [Vector]? & 1 & 3 & 0 \\
    \midrule
    What vector transmits a pathogen which causes a disease which is treated by [Medication 1] and [Medication 2]? [if one] & 0 & 3 & 1 \\
    What medication treats a disease which is caused by a pathogen which is transmitted by [Vector 1] and [Vector 2]? [if one] & 0 & 3 & 1 \\
    \midrule
    What vectors transmit a pathogen which causes a disease which is treated by [Medication 1] and [Medication 2]? & 1 & 3 & 1 \\
    What medications treat a disease which is caused by a pathogen which is transmitted by [Vector 1] and [Vector 2]? & 1 & 3 & 1 \\
    \midrule
    What vector transmits a pathogen which causes a disease which is treated by [Medication 1] but not [Medication 2]? [if one] & 0 & 3 & 2 \\
    What medication treats a disease which is caused by a pathogen which is transmitted by [Vector 1] but not [Vector 2]? [if one] & 0 & 3 & 2 \\
    \midrule
    What vectors transmit a pathogen which causes a disease which is treated by [Medication 1] but not [Medication 2]? & 1 & 3 & 2 \\
    What medications treat a disease which is caused by a pathogen which is transmitted by [Vector 1] but not [Vector 2]? & 1 & 3 & 2 \\
    \bottomrule
    \end{tabular}
    }
    \label{tab:ablation_templatex_L3}
\end{table*}

\clearpage

\section{Dataset Release and Leaderboard Setup}
\label{app:dataset_release}

To facilitate reproducibility and future research \citep{tatarinov2025languagemodelingfuturefinance}, we are publicly releasing (under CC-BY NC ND 4.0 license) QA pairs for 40 documents out of the total 170 as the development set, allowing researchers to validate model performance under controlled conditions. The remaining QA pairs serve as the test set, which is not publicly available to prevent data contamination \citep{sainz-etal-2023-nlp}. For this test set, we are making the questions public, without the ground-truth answers. The distribution of the test and dev sets is provided in Table \ref{tab:qa_pairs_stats}. In addition, we are hosting a leaderboard\footnote{\url{https://huggingface.co/spaces/gtfintechlab/KG-MuLQA-D-Leaderboard}}, following standard practices in recent LLM evaluation benchmarks \citep{10.1145/3711896.3737417,yue2024mmmu,zhao-etal-2024-knowledgefmath,lu2023mathvista}. 

\begin{table}[!htbp]
    \caption{Dataset statistics of the constructed \datasetshortname{} dataset.}
    \label{tab:qa_pairs_stats}
    \renewcommand{\arraystretch}{1.2}  
    \centering
    \footnotesize
    \resizebox{\linewidth}{!}{%
    \begin{tabular}{lccc}
        \toprule
        \textbf{Stats} & \textbf{Dev} & \textbf{Test} & \textbf{Total} \\
        \midrule
        \# Documents & 40 & 130 & 170 \\
        \cdashline{1-4}[2pt/2pt]
        \# Question per doc (min) & 1 & 1 & 1 \\
        \# Question per doc (avg) & 14.75 & 23.49 & 21.44 \\
        \# Question per doc (max) & 83 & 428 & 428 \\
        \cdashline{1-4}[2pt/2pt]
        \# Easy Questions (count) & 1,499 & 5,051 & 6,550 \\
        \# Medium Questions (count) & 2,680 & 10,203 & 12,883 \\
        \# Hard Questions (count) & 239 & 467 & 706 \\
        \cdashline{1-4}[2pt/2pt]
        \textbf{\# Questions (count)} & 4,418 & 15,721 & 20,139 \\
        \bottomrule
    \end{tabular}
    }
\end{table}

\section{Author Contribution}

NT led the data acquisition by extracting credit agreements from SEC EDGAR, developed the annotation guidelines specifying the entity and relation schemas, conceptualized the RDF-based knowledge graph framework, devised multi‑dimensional QA templates, and benchmarked long‑context LLMs in both RAG and oracle settings. 
VK extensively annotated and cleaned documents, identified edge cases and was involved in developing the KG construction and question-answer generation modules. She performed error analysis, led human evaluation, and constructed visualizations. 
HS contributed to the document annotation process, developed the pipelines for KG construction and SPARQL QA generation and assisted in LLM response analysis.
AR implemented the initial end‑to‑end benchmarking pipeline and supported the human evaluation. He experimented with various prompting techniques and evaluation metrics and created one-shot prompts used in experiments.
HSA contributed to all annotation stages and SPARQL‑based question generation. 
VS, AL, and RL contributed to annotation, data cleaning, and manuscript refinement. 
AS supervised all iterations of the annotation process, guided edge‑case incorporation, shaped research direction for document-level question answering, advised on ablation design, and contributed to error analysis and writing.
SC provided overall project supervision and guidance. 

\section{Acknowledgments}

We would like to thank Anvita Mahajan, Archishman VB, Arvind K R, Daksh Jain, Deng Li, Jorge Navarro Gracia, Kaustubh Gayadhankar, Krish Shah, Kritika Bansal, Meher Bhardwaj, Nilesh Gupta, Piyush Mohapatra, Pranav Krishna, Rushi Glasswala, Shashwat Bajpai, Soumyajit Basu, Tvisha Shah, Vishnu Varma, Xu Huang and Udish Jangid for contributing to the initial round of annotations and early-stage coding experiments.

\end{document}